\documentclass{article}

\PassOptionsToPackage{numbers, compress}{natbib}

\usepackage[preprint]{neurips_2026}
\usepackage[most]{tcolorbox}

\usepackage{amsmath}
\usepackage{amssymb}
\usepackage{listings}
\usepackage{xcolor}
\usepackage{comment}
\usepackage{url}
\usepackage{subcaption}
\usepackage{diagbox}

\usepackage{graphicx} 
\usepackage{enumitem}
\usepackage{tikz}
\usetikzlibrary{positioning,arrows.meta,calc,fit,shapes.geometric,backgrounds}
\usepackage{fancyhdr}
\usepackage{graphicx}
\usepackage[ruled,vlined,linesnumbered]{algorithm2e}
\SetAlCapSkip{0.5em}
\SetNlSty{textbf}{}{:}
\usepackage{booktabs}

\usetikzlibrary{positioning, calc, decorations.pathreplacing, arrows.meta}

\definecolor{codegreen}{rgb}{0,0.6,0}
\definecolor{codegray}{rgb}{0.5,0.5,0.5}
\definecolor{codepurple}{rgb}{0.58,0,0.82}
\definecolor{backcolour}{rgb}{0.95,0.95,0.92}

\lstdefinestyle{python}{
    backgroundcolor=\color{backcolour},   
    commentstyle=\color{codegreen},
    keywordstyle=\color{magenta},
    numberstyle=\tiny\color{codegray},
    stringstyle=\color{codepurple},
    basicstyle=\ttfamily\footnotesize,
    breakatwhitespace=false,         
    breaklines=true,                 
    captionpos=b,                    
    keepspaces=true,                 
    numbers=left,                    
    numbersep=5pt,                  
    showspaces=false,                
    showstringspaces=false,
    showtabs=false,                  
    tabsize=2
}

\newtcolorbox{perspective}{
    colback=blue!5!white,
    colframe=blue!75!black,
    fonttitle=\bfseries,
    title=Perspective,
    arc=4pt 
}

\title{Long Context Pre-Training with Lighthouse Attention}
\author{%
  Bowen~Peng\textsuperscript{\textasteriskcentered} \\
  Nous~Research \\
  \texttt{bloc@nousresearch.com} \\
  \And
  Subho~Ghosh\textsuperscript{\textasteriskcentered} \\
  Nous~Research \\
  \texttt{subho@nousresearch.com} \\
  \And
  Jeffrey~Quesnelle \\
  Nous~Research \\
  \texttt{emozilla@nousresearch.com} \\
}

\date{February 2026}

\begin{document}

\maketitle

\renewcommand{\thefootnote}{\fnsymbol{footnote}}
\footnotetext[1]{Equal contribution.}
\renewcommand{\thefootnote}{\arabic{footnote}}

\begin{abstract}
   Training causal transformers at extreme sequence lengths is bottlenecked by the quadratic time and memory of scaled dot-product attention (SDPA). In this work, we propose Lighthouse Attention, a training-only symmetrical selection-based hierarchical attention algorithm that wraps around ordinary SDPA and can be easily removed towards the end of the training. Our hierarchical selection is also gradient-free, which exempts us from dealing with a complicated and potentially inefficient backward pass kernel. Our contribution is three-fold: (i) A subquadratic hierarchical pre- and post-processing step that does adaptive compression and decompression of the sequence. (ii) A symmetrical compression strategy that pools queries, keys and values at the same time, while preserving left-to-right causality, which greatly improves parallelism. (iii) A two stage training approach which we pre-train for the majority of the time with Lighthouse Attention and recover a full attention model at the end with a short training. We run preliminary small scale LLM pre-training experiments that show the effectiveness of our method compared to full attention training with all other settings matched, where we achieve a faster total training time and lower final loss after the recovery phase. \\ \\ Full code is available at: \\ \url{https://github.com/ighoshsubho/lighthouse-attention}.

\end{abstract}

\section{Introduction}          
  The frontier of language modeling has moved toward contexts of 128K, 1M, and longer, pushed by agentic multi-step reasoning, long-document understanding, and interleaved multimodal inputs~\citep{openai2024gpt4,anthropic2024claude3,geminiteam2024gemini,meta2024llama3,qwen2025qwen25,deepseek2024v3,moonshotai2025kimi}. Training at these scales is the dominant hardware bottleneck: scaled dot-product attention has $\Theta(N^2)$ compute and memory, a wall that FlashAttention~\citep{shah2024flashattention3} pushes back but does not remove.
  
  A growing body of work replaces dense attention with selection: each query attends only to a small subset of keys. Block-level methods such as MoBA~\citep{lu2025moba} and Native Sparse 
  Attention~\citep{yuan2025nsa} select contiguous blocks, while token-level methods such as DeepSeek Sparse Attention (DSA;~\citealp{deepseek2025dsa}) score every past token via a learned indexer and forward the top-$k$ into a sparse attention operator; HISA~\citep{zhao2026hisa} adds a hierarchical indexer to keep scoring from becoming the new bottleneck. These methods 
  produce meaningful inference speedups but inherit two design decisions that fit long-context pretraining poorly. \emph{(i) Asymmetry}: queries stay at full resolution while keys and
  values are pooled, so the hierarchy serves only as a compressed addressable memory rather than a multi-scale representation. \emph{(ii) Architectural entanglement}: selection lives
  inside the attention kernel, so the carefully optimized dense-attention kernels that modern tensor-core GPUs accelerate cannot be reused; every sparse method ships its own kernel.

  There is also a concern specific to training. An inference-time sparse method~\citep{zhao2026hisa,ribar2024sparq,tang2024quest,zhang2023h2o,xiao2024streamingllm} is by construction as good as its dense backbone, since the sparse substitution is evaluated only against the dense forward. A training-time sparse method must survive a harder test: once training is done, will the resulting model still be a competent dense-attention model? 
  
  We take this last question as our central correctness criterion. We introduce Lighthouse Attention: a selection-based hierarchical attention that pools $Q, K, V$ \emph{symmetrically} across a multi-level pyramid, scores every pyramid entry bidirectionally with a parameter-free scorer, and selects the top-$K$ entries with a fused chunked-bitonic kernel. The selected entries
  form a dense, causally consistent sub-sequence attended to with stock FlashAttention; outputs are scattered back through a deterministic kernel. The top-$K$ step is non-differentiable, 
  with no straight-through estimator: gradients flow through scatter, FlashAttention, and gather into $W_Q, W_K, W_V$, which learn to produce values that are useful when selected. No
  auxiliary parameters or losses are added. Two consequences follow: the symmetric pyramid is a full multi-scale representation rather than a compressed context, and because selection
  sits outside the attention path, the expensive step is stock FlashAttention on a sub-sequence of size $\mathcal{O}(L p K + N/p^{L-1})$, which reduces to $\mathcal{O}(N \log N)$ at $L =
  \log_p(N/K)$.

  Our central empirical finding addresses the training-correctness concern directly: after a brief dense-SDPA resumption, Lighthouse-trained models match or beat a fully dense-SDPA       
  baseline trained from scratch on the same token budget. The hierarchical training signal does not hollow out the model's ability to use full attention at inference, a property inference-only sparse methods cannot claim because they never touch the training loop. 
  We summarize our contributions:                                                                                  
  \begin{itemize}[noitemsep, topsep=0pt]
    \item A selection-based hierarchical attention designed for long-context pretraining with symmetric $Q/K/V$ pooling, bidirectional top-$K$ selection, and stock FlashAttention on the gathered sub-sequence, keeping sparse logic entirely outside the attention kernel.
    \item Fused GPU kernels (chunked-bitonic top-$K$ and a custom scatter-back) that make this design fast at very large contexts.
    \item The strongest empirical criterion for a training-time hierarchical method to our knowledge: dense-SDPA resumption after Lighthouse pretraining matches a dense-from-scratch baseline on
   training loss.                                                                                                                                                                          
  \end{itemize}
    
\section{Related works}
    \paragraph{Compression and pruning.} A first response to quadratic attention abandons softmax for a bounded-size state: linear attention~\citep{katharopoulos2020transformers,choromanski2021performer}, state-space and gated variants~\citep{gu2023mamba,gu2024mamba2,yang2024gla,sun2023retentive}, and log-linear attention~\citep{guo2025log}: which gives strong asymptotics but compresses the entire past and limits long-range recall~\citep{arora2023zoology}. A second keeps softmax and prunes at block granularity, either training-free (MInference, FlexPrefill, XAttention, SpargeAttention~\citep{jiang2024minference,lai2025flexprefill,xu2025xattention,zhang2025sparge}) or end-to-end (MoBA, NSA~\citep{lu2025moba,yuan2025nsa}); these map cleanly onto tiled matmul but force a single retain/discard decision per block and pool only the key--value side. A third prunes at token granularity, mostly at inference for KV-cache eviction (H2O, TOVA, SnapKV, LazyLLM, Quest, SparQ~\citep{zhang2024h2o,oren2024tova,li2024snapkv,fu2024lazyllm,tang2024quest,ribar2024sparq}), or via a learned indexer trained end-to-end (DSA~\citep{deepseek2025dsa}). The defining property of this family is that once selection is identified it is welded into the attention operator as a custom sparse matmul or per-query gather, foreclosing reuse of stock dense kernels.
    
    \paragraph{Hierarchies and training-time correctness.} Multi-resolution attention~\citep{yang2016hierarchical} has returned to sparse LLM attention in two flavors. NSA~\citep{yuan2025nsa}, InfLLM-V2~\citep{zhao2026infllmv2}, Twilight~\citep{lin2025twilight}, and DoubleP~\citep{ni2026doublep} build hierarchies that the attention itself reads from compression branches, centroid summaries, or quantized proxies. HISA~\citep{zhao2026hisa} is a training-free, plug-in replacement for DSA's indexer that runs a block-to-token two-stage score and forwards the selected tokens unchanged to the same Sparse MLA operator DSA already uses. In every case the hierarchy applies only to keys and values, and the selection that emerges still feeds a custom sparse attention kernel. Lighthouse differs on three axes: it pools queries symmetrically with keys and values into coherent multi-resolution $(Q^{(\ell)},K^{(\ell)},V^{(\ell)})$ triples; the pyramid is used purely to rank and select, so the attention that follows is stock FlashAttention on a dense sub-sequence with no sparse indexing inside the kernel; and it is trained end-to-end through a non-differentiable top-$k$ wrapped by a differentiable scatter, with no auxiliary loss or straight-through estimator. Inference-only sparse methods (including HISA) inherit a correctness floor from their underlying dense model, but training-time sparse methods (MoBA, NSA) must answer whether the weights they produce remain competent dense models. We take a brief dense-SDPA resumption recovering the quality of a dense-from-scratch baseline as our central correctness criterion.

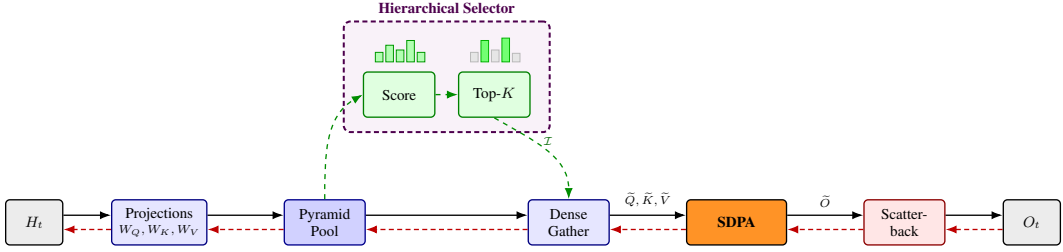
\begin{figure}[t]
\centering
\resizebox{\linewidth}{!}{%
    \begin{tikzpicture}[                                         >={Latex[length=2mm,width=1.5mm]},                           every node/.style={font=\small},                             fwd/.style       ={-Latex, thick},
    selflow/.style   ={-Latex, thick, green!55!black, dashed},   grad/.style      ={-Latex, thick, red!75!black, dash pattern=on 3pt off 2pt},                                     box/.style       ={rectangle, rounded corners=2.5pt, draw, thick,                                                       align=center, minimum height=0.95cm, inner sep=3pt},         inbox/.style     ={box, fill=gray!15,  minimum width=1.2cm}, projbox/.style   ={box, fill=blue!10,  draw=blue!50!black, minimum width=2.0cm},                                        pyrbox/.style    ={box, fill=blue!18,  draw=blue!55!black, minimum width=1.7cm},                                        selbox/.style    ={box, fill=green!12, draw=green!50!black, minimum width=1.5cm},                                        gatherbox/.style ={box, fill=blue!10,  draw=blue!50!black, minimum width=1.7cm},                                        flashbox/.style  ={box, fill=orange!85, draw=black, text=black, minimum width=2.1cm, font=\small\bfseries},      scatterbox/.style={box, fill=red!10,   draw=red!55!black, minimum width=1.7cm},                                        outbox/.style    ={box, fill=gray!15,  minimum width=1.2cm}, bar/.style       ={draw=green!60!black, fill=green!30, line width=0.3pt},                                                barsel/.style    ={draw=green!70!black, fill=green!55, line width=0.3pt},                                                bargray/.style   ={draw=gray!55,        fill=gray!20, line width=0.3pt},
    ]                                                            
    \node[inbox]                          (H)    {$H_t$};
    \node[projbox,    right=10mm of H]    (proj) {Projections\\[-1pt]\scriptsize $W_Q,W_K,W_V$};                                               
    \node[pyrbox,     right=16mm of proj] (pyr)  {Pyramid\\Pool};                                                                              
    \node[gatherbox,  right=34mm of pyr]  (gat)  {Dense\\Gather};                                                                              
    \node[flashbox,   right=16mm of gat]  (fa)   {SDPA};                                                                          
    \node[scatterbox, right=16mm of fa]   (sc)   {Scatter-\\back};                                                                             
    \node[outbox,     right=12mm of sc]   (O)    {$O_t$};         
    \coordinate (mid) at ($(pyr.east)!0.5!(gat.west)$);
    \node[selbox, above=22mm of mid, xshift=-10mm] (score) {Score};                                                                            
    \node[selbox, above=22mm of mid, xshift= 10mm] (tk)    {Top-$K$};                                                                          
    \draw[selflow] (score) -- (tk);                              
    \path (score.north) ++(0,2mm) coordinate (si);
    \foreach \x/\h in {0/2.0, 1/3.5, 2/2.5, 3/4.5, 4/2.0}                                                                                      
      \draw[bar] ($(si)+(-5mm+\x*2.2mm,0)$) rectangle ++(1.6mm,{\h*1mm});                                                                      
    \coordinate (scoreIconTop) at ($(si)+(0,5mm)$);                                                                           \path (tk.north) ++(0,2mm) coordinate (ti);                                                                                                
    \draw[bargray] ($(ti)+(-5.0mm,0)$) rectangle ++(1.6mm,2.0mm);                                                                              
    \draw[barsel]  ($(ti)+(-2.8mm,0)$) rectangle ++(1.6mm,4.5mm);                                                                              
    \draw[bargray] ($(ti)+(-0.6mm,0)$) rectangle ++(1.6mm,2.5mm);                                                                              
    \draw[barsel]  ($(ti)+( 1.6mm,0)$) rectangle ++(1.6mm,5.0mm);                                                                              
    \draw[bargray] ($(ti)+( 3.8mm,0)$) rectangle ++(1.6mm,1.8mm);                                                                              
    \coordinate (tkIconTop) at ($(ti)+(0,5.5mm)$);               
    \begin{scope}[on background layer]
      \node[draw=violet!60!black, very thick, dash pattern=on 3pt off 2pt,                                                                     
            rounded corners=4pt, fill=violet!5,                                                                                                
            fit=(score)(tk)(scoreIconTop)(tkIconTop),
            inner xsep=4mm, inner ysep=3mm,                                                                                                    
            label={[font=\small\bfseries, violet!60!black, yshift=1pt] above:Hierarchical Selector}]
            (selgrp) {};                                                                                                                       
    \end{scope}   
    \draw[fwd] ($(H.east)    +(0,1.5mm)$) -- ($(proj.west)+(0,1.5mm)$);
    \draw[fwd] ($(proj.east) +(0,1.5mm)$) -- ($(pyr.west) +(0,1.5mm)$);                                                                        
    \draw[fwd] ($(pyr.east)  +(0,1.5mm)$) -- ($(gat.west) +(0,1.5mm)$);                                                                        
    \draw[fwd] ($(gat.east)  +(0,1.5mm)$) -- node[above=0.5mm, font=\scriptsize] {$\widetilde Q,\widetilde K,\widetilde V$} ($(fa.west)        
  +(0,1.5mm)$);                                                                                                                                
    \draw[fwd] ($(fa.east)   +(0,1.5mm)$) -- node[above=0.5mm, font=\scriptsize] {$\widetilde O$} ($(sc.west)  +(0,1.5mm)$);                   
    \draw[fwd] ($(sc.east)   +(0,1.5mm)$) -- ($(O.west)   +(0,1.5mm)$);                                                 
    \draw[selflow] (pyr.north) to[out=90,  in=210] (score.west);
    \draw[selflow] (tk.south)  to[out=-30, in=90]  (gat.north);                                                                                
    \node[font=\scriptsize, green!40!black]
          at ($(tk.south)!0.4!(gat.north)+(5mm,2mm)$) {$\mathcal{I}$};                                                                         
                  
    \draw[grad] ($(O.west)    +(0,-1.5mm)$) -- ($(sc.east)  +(0,-1.5mm)$);                                                                     
    \draw[grad] ($(sc.west)   +(0,-1.5mm)$) -- ($(fa.east)  +(0,-1.5mm)$);                                                                     
    \draw[grad] ($(fa.west)   +(0,-1.5mm)$) -- ($(gat.east) +(0,-1.5mm)$);                                                                     
    \draw[grad] ($(gat.west)  +(0,-1.5mm)$) -- ($(pyr.east) +(0,-1.5mm)$);                                                                     
    \draw[grad] ($(pyr.west)  +(0,-1.5mm)$) -- ($(proj.east)+(0,-1.5mm)$);                                                                     
    \draw[grad] ($(proj.west) +(0,-1.5mm)$) -- ($(H.east)   +(0,-1.5mm)$);                                                                    
    \end{tikzpicture}}
  \caption{\textbf{Lighthouse Attention architecture.} \textit{Forward} (black): $H_t$ is projected to $Q,K,V$, passed through the symmetric \emph{Pyramid Pool} on the trunk, and guided by indices $\mathcal{I}$ from the \emph{Hierarchical Selector}, is fed to a dense gather which topographically sorts the gathered hierarchies into a single contiguous and causal sequence, then stock SDPA, and scatter-back to produce $O_t$. \textit{Selection} (green): the selector taps the pooled summaries off the trunk; the parameter-free scorer ranks them by $\ell_2$ norm and a top-$K$ kernel keeps the largest entries, emitting integer indices $\mathcal{I}$ that merge back into the trunk at the dense gather. \textit{Gradient} (red, dashed): $\nabla L$ travels along the trunk ($O_t \!\to\!$ scatter $\!\to\!$ FA $\!\to\!$ gather $\!\to\!$ pyramid pool $\!\to\! W_{Q,K,V} \!\to\! H_t$); the selector branch is non-differentiable and is bypassed.}                                                  \label{fig:pipeline}                                           \end{figure} 

\section{Method}
\label{sec:method}

We present Lighthouse Attention, a selection-based hierarchical attention mechanism for long-context pretraining. Lighthouse replaces a standard Transformer attention layer with a four-stage pipeline that surrounds, but does not modify, the attention kernel: a pre-attention selection stage drives a contiguous gather, stock FlashAttention~\citep{dao2023flashattention2} runs on the gathered sub-sequence, and a post-attention scatter writes the result back to the original positions. Selection is driven by a parameter-free scoring functional over a multi-resolution pyramid of the layer's own queries, keys, and values, so Lighthouse introduces \emph{no new learnable parameters} beyond those of the underlying attention block.

\subsection{Preliminaries}
  Let $X \in \mathbb{R}^{N \times d_{\text{model}}}$ be the input, $W_Q, W_K, W_V \in \mathbb{R}^{d_{\text{model}} \times d}$ projection matrices for one head, and $M \in \mathbb{R}^{N
  \times N}$ a causal mask. Standard scaled dot-product attention~\citep{dao2022flashattention} is                                                                                         
  \begin{equation}
    Q = X W_Q,\quad K = X W_K,\quad V = X W_V,\qquad \mathrm{Attn}(Q, K, V) = \mathrm{softmax}\!\left(\tfrac{Q K^\top}{\sqrt{d}} + M\right) V,                                               
    \label{eq:sdpa}
  \end{equation}
  with both time and memory cost $\Theta(N^2 d)$. FlashAttention reduces constants but not asymptotics; at $N \ge 10^5$ this term dominates. Lighthouse replaces Eq.~\eqref{eq:sdpa} with: \textbf{(i)} symmetric average-pooling of $Q,K,V$ into an $L$-level pyramid (factor $p$); \textbf{(ii)} parameter-free scoring and a fused 
  chunked-bitonic top-$k$ selection over all levels jointly; \textbf{(iii)} stock FlashAttention on a contiguous sub-sequence of $S \ll N$ selected entries; \textbf{(iv)} a scatter-back that distributes each output to the $p^\ell$ base positions it represents. Stages (ii) and (iv) are custom kernels (Sec.~\ref{sec:kernels}); stage (iii) is the same FlashAttention call 
  as the dense baseline. The top-$k$ is treated as discrete and non-differentiable: indices carry no gradient and the scoring functional is not trained. Gradients reach $W_Q, W_K, W_V$ only through stages (iv), (iii), and the gather: the projections learn to produce values that are \emph{useful when selected} rather than scores that are good at selecting, sidestepping the optimization fragility of learnable selectors.

\begin{figure}[t]                      
  \centering
  \resizebox{\linewidth}{!}{%
  \begin{tikzpicture}[                                                                                                     
      >={Latex[length=2.5mm,width=2mm]},                                                                                   
      every node/.style={font=\small},                                                                                     
      fwd/.style       ={-Latex, thick, black!75},
      cardP/.style     ={rectangle, rounded corners=4pt, draw=blue!55!black, very thick,                                   
                           fill=blue!4, minimum width=6cm, minimum height=4.6cm, align=center, inner sep=0pt},             
      cardS/.style     ={rectangle, rounded corners=4pt, draw=green!55!black, very thick,                                  
                           fill=green!4, minimum width=6cm, minimum height=4.6cm, align=center, inner sep=0pt},            
      cardT/.style     ={rectangle, rounded corners=4pt, draw=violet!55!black, very thick,                                 
                           fill=violet!4, minimum width=6.5cm, minimum height=4.6cm, align=center, inner sep=0pt},         
      pcell/.style     ={draw=blue!60!black, fill=blue!25, line width=0.3pt, inner sep=0pt},                               
      bar/.style       ={draw=green!65!black, fill=green!35, line width=0.3pt},                                            
      barsel/.style    ={draw=violet!70!black, fill=violet!55, line width=0.4pt},                                          
      barkept/.style   ={draw=violet!55!black, fill=violet!22, line width=0.3pt},                                          
      poolline/.style  ={draw=blue!50!black, line width=0.6pt},                                                            
      expline/.style   ={draw=violet!55!black, line width=0.6pt},                                                          
      note/.style      ={font=\scriptsize, gray!45!black, align=center},                                                   
      ttl/.style       ={font=\bfseries\small},                                                                            
  ]                                                                                                                        
                  
  \node[cardP] (P) at (0,0)            {};                                                                                 
  \node[cardS, right=28mm of P] (S)    {};
  \node[cardT, right=22mm of S] (T)    {};                               
  \node[ttl, text=blue!55!black, anchor=north] at ($(P.north)+(0,-3mm)$) {1. Pyramid Pool};
  \foreach \i in {0,1}                                                                                                     
    \node[pcell, minimum width=11mm, minimum height=3mm] (P2-\i) at ($(P.center)+(\i*16mm - 8mm, 9mm)$) {};
  \foreach \i in {0,...,3}                                                                                                 
    \node[pcell, minimum width=4.5mm, minimum height=3mm] (P1-\i) at ($(P.center)+(\i*8mm - 12mm, 1mm)$) {};               
  \foreach \i in {0,...,7}                                                                                                 
    \node[pcell, minimum width=1.8mm, minimum height=3mm] (P0-\i) at ($(P.center)+(\i*4mm - 14mm, -7mm)$) {};              
  \foreach \p/\ca/\cb in {0/0/1, 1/2/3} {                                                                                  
      \draw[poolline] (P1-\ca.north) -- (P2-\p.south);                                                                     
      \draw[poolline] (P1-\cb.north) -- (P2-\p.south);                                                                     
  }                                                                                                                        
  \foreach \p/\ca/\cb in {0/0/1, 1/2/3, 2/4/5, 3/6/7} {                                                                    
      \draw[poolline] (P0-\ca.north) -- (P1-\p.south);                                                                     
      \draw[poolline] (P0-\cb.north) -- (P1-\p.south);
  }                                                                                                                        
  \node[note, anchor=east] at ($(P.center)+(-15mm, 9mm)$) {$\ell{=}2$};
  \node[note, anchor=east] at ($(P.center)+(-15mm, 1mm)$) {$\ell{=}1$};                                                    
  \node[note, anchor=east] at ($(P.center)+(-15mm,-7mm)$) {$\ell{=}0$};
  \node[note, anchor=south, align=center] at ($(P.south)+(0,3mm)$) {mean-pool by $p^\ell$};                                
  \node[ttl, text=green!50!black, anchor=north] at ($(S.north)+(0,-3mm)$) {2. Norm Score};
  \foreach \i/\h in {0/3,1/5,2/2,3/4,4/6,5/3,6/4,7/2}                                                                      
    \draw[bar] ($(S.center)+(-16mm + \i*4mm, -7mm)$) rectangle ++(3mm, {\h*0.7mm});                                        
  \foreach \i/\h in {0/5,1/4,2/6,3/4}                                                                                      
    \draw[bar] ($(S.center)+(-16mm + \i*8mm, 1mm)$) rectangle ++(7mm, {\h*0.7mm});                                         
  \foreach \i/\h in {0/5,1/6}                                                                                              
    \draw[bar] ($(S.center)+(-16mm + \i*16mm, 9mm)$) rectangle ++(15mm, {\h*0.7mm});                                       
  \node[note, anchor=east] at ($(S.center)+(-18mm, -5.5mm)$) {$\ell{=}0$};                                                 
  \node[note, anchor=east] at ($(S.center)+(-18mm,  2.5mm)$) {$\ell{=}1$};                                                 
  \node[note, anchor=east] at ($(S.center)+(-18mm, 10.5mm)$) {$\ell{=}2$};                                                 
  \node[note, anchor=south, align=center] at ($(S.south)+(0,3mm)$) {$\ell_2$ norm + max-pool};                             
  \node[ttl, text=violet!55!black, anchor=north]
       at ($(T.north)+(0,-3mm)$) {3. Chunked Bitonic Top-$K$};                                                             
  \foreach \i/\sty/\h in {0/barkept/3, 1/barsel/6, 2/barkept/2, 3/barsel/5}                                                
    \node[\sty, minimum width=7mm, minimum height={\h*1mm}] (TL-\i)                                                        
       at ($(T.center)+(-16mm + \i*8mm + 3.5mm, 9mm)+(0,{\h*0.5mm})$) {};                                                  
  \foreach \i/\sty/\h in {0/barsel/5, 1/barkept/3, 2/barkept/2, 3/barsel/4}                                                
    \node[\sty, minimum width=7mm, minimum height={\h*1mm}] (TM-\i)                                                        
       at ($(T.center)+(-16mm + \i*8mm + 3.5mm, 1mm)+(0,{\h*0.5mm})$) {};                                                  
  \draw[expline, ->, dashed] (TL-1.south) -- ($(TM-0.north)+(-1mm,0)$);                                                    
  \draw[expline, ->, dashed] (TL-1.south) -- ($(TM-1.north)+(-1mm,0)$);                                                    
  \draw[expline, ->, dashed] (TL-3.south) -- ($(TM-2.north)+(-1mm,0)$);                                                    
  \draw[expline, ->, dashed] (TL-3.south) -- ($(TM-3.north)+(-1mm,0)$);                                                    
  \foreach \i in {0,...,3}                                                                                                 
    \node[barsel, minimum width=7mm, minimum height=4mm] (TF-\i)                                                           
       at ($(T.center)+(-16mm + \i*8mm + 3.5mm, -7mm)+(0,2mm)$) {};                                                        
  \draw[expline, ->, dashed] (TM-0.south) -- ($(TF-0.north)+(-1mm,0)$);
  \draw[expline, ->, dashed] (TM-0.south) -- ($(TF-1.north)+(-1mm,0)$);                                                    
  \draw[expline, ->, dashed] (TM-3.south) -- ($(TF-2.north)+(-1mm,0)$);                                                    
  \draw[expline, ->, dashed] (TM-3.south) -- ($(TF-3.north)+(-1mm,0)$);                                                    
  \node[note, anchor=east] at ($(T.center)+(-18mm, 11mm)$)  {$\ell{=}2$};                                                  
  \node[note, anchor=east] at ($(T.center)+(-18mm,  2.5mm)$){$\ell{=}1$};                                                  
  \node[note, anchor=east] at ($(T.center)+(-18mm, -5mm)$)  {$\ell{=}0$};                                                  
  \node[note, anchor=south, align=center] at ($(T.south)+(0,3mm)$)                                                         
     {\tikz\node[barsel, minimum width=2.5mm, minimum height=1.8mm]{};\,parent\quad                                        
      \tikz\node[barkept,minimum width=2.5mm, minimum height=1.8mm]{};\,kept\\                                             
      bitonic argsort + tree-prune};
  \node[font=\small, align=center, anchor=east] (in)
       at ($(P.west)+(-12mm,0)$) {$\mathbf{Q,K,V}$\\[-1pt]\scriptsize $[B,S,H,D]$};
  \node[font=\small, align=center, anchor=west] (out)                                                                      
       at ($(T.east)+(12mm,0)$) {$\mathcal{I}$\\[-1pt]\scriptsize $[B,H,K]$};                                              
  \draw[fwd] (in.east) -- (P.west);                                                                                        
  \draw[fwd] (P.east)  -- node[above=2pt, font=\scriptsize, align=center] {pyramid\\tokens} (S.west);                      
  \draw[fwd] (S.east)  -- node[above=2pt, font=\scriptsize, align=center]                                                  
              {scores$_{qk}$\\scores$_{kq}$} (T.west);                                                                     
  \draw[fwd] (T.east)  -- (out.west);                                    
  \begin{scope}[on background layer]
    \node[draw=violet!60!black, very thick, dash pattern=on 4pt off 2pt,                                                   
          rounded corners=5pt, fill=violet!2,
          fit=(S)(T), inner xsep=4mm, inner ysep=9mm,                                                                      
          label={[font=\bfseries, text=violet!60!black, yshift=2pt] above:Hierarchical Selector}] {};
  \end{scope}                                                                                                              
  \end{tikzpicture}}
  \caption{\textbf{Pyramid Pool and the Hierarchical Selector.} The Pyramid Pool is a fixed pre-selection stage that lives 
  \emph{outside} the selector. \textbf{(1) Pyramid Pool} mean-pools $Q,K,V$ by $p^\ell$; lines show which tokens feed each 
  summary. The pooled tokens enter the selector, where \textbf{(2) Norm Score} computes parameter-free $\ell_2$ norms
  $\|Q^{(\ell)}\|_2,\|K^{(\ell)}\|_2$ (coarser levels reuse finer norms via max-pool) and \textbf{(3) Chunked Bitonic      
  Top-$K$} keeps top-$K$ \emph{parents} (dark) that descend, while rejected-kept entries (light) emit to $\mathcal{I}$
  without descending. Pruning is implicit: children of non-parents are never expanded. Per-tile bitonic argsort runs
  in-register; uint64 packed keys let a single \texttt{torch.sort} produce $\mathcal{I}$.}
  \label{fig:selector-internals}
  \end{figure}
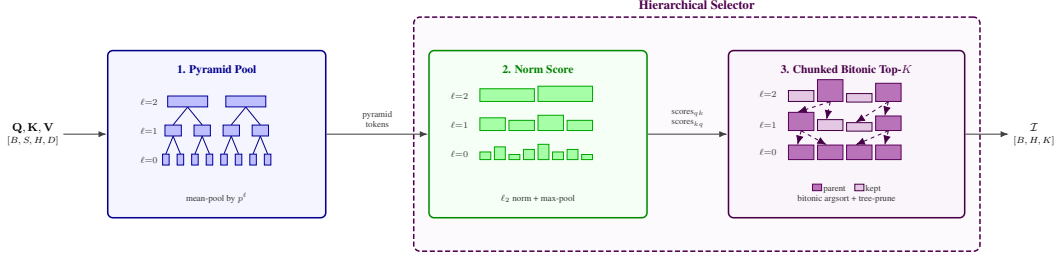 

\subsection{Overview}
\label{sec:method-overview}
A Lighthouse attention layer replaces standard scaled dot-product attention (Eq.~\eqref{eq:sdpa}) with a four-stage pipeline that surrounds, but does not modify, the attention kernel. Let $Q, K, V \in \mathbb{R}^{N \times d}$ be the per-head projections from the layer's own $W_Q, W_K, W_V$ (Sec.~3.1).
\begin{enumerate}[noitemsep, topsep=2pt, leftmargin=*]
    \item[\textbf{(i)}] \textbf{Pyramid.} Average-pool $Q, K, V$ symmetrically into an $L$-level pyramid with pooling factor $p$, producing coherent triples $(Q^{(\ell)}, K^{(\ell)}, V^{(\ell)})$ for $\ell = 0, \dots, L-1$.
    \item[\textbf{(ii)}] \textbf{Score and top-$k$.} Assign each pyramid entry parameter-free query and key scores and select the $k$ entries with the highest combined relevance across   
  all levels via a fused chunked-bitonic top-$k$ kernel.
    \item[\textbf{(iii)}] \textbf{Dense sub-sequence attention.} Gather the selected triples into a contiguous sub-sequence of length $S$ and compute softmax attention over it with stock FlashAttention.
    \item[\textbf{(iv)}] \textbf{Scatter-back.} Distribute each entry's output to the $p^{\ell}$ base positions it represents via a deterministic integer-atomic scatter kernel.
 \end{enumerate}                                                                                                                                                                          
  Stages (ii) and (iv) are custom kernel (Sec.~\ref{sec:kernels}); stage (iii) is the same FlashAttention call used by the dense baseline. Lighthouse adds no learnable parameters or losses: the pyramid is a fixed pooling, the scorer is parameter-free, and gather/scatter are data-flow primitives. Gradients flow from the loss through stages~(iv) and~(iii) into the gathered 
  $Q, K, V$ and on into $W_Q, W_K, W_V$; the top-$k$ step is discrete and non-differentiable, so its indices carry no gradient and we use no straight-through estimator. The projections
  therefore learn to be useful \emph{when selected}, not to score well at selecting.

\subsection{Pyramid Construction}
\label{sec:method-pyramid}

Given $Q, K, V \in \mathbb{R}^{N \times d}$, Lighthouse Attention constructs an $L$-level pyramid whose $\ell$-th level is a non-overlapping window pooling of the previous level. For $\ell = 0, \dots, L-1$, define the $i$-th window at level $\ell$ as
\begin{equation}                                                             \mathcal{W}^{(\ell)}_i \;=\; \bigl[\, i\, p^{\ell},\; (i+1)\, p^{\ell} - 1 \,\bigr],                                                                 \qquad i = 0, \dots, \tfrac{N}{p^{\ell}} - 1,                         \end{equation}                                                            where $p$ is the pooling factor. The pyramid entries are then             \begin{equation}                                                             Q^{(\ell)}_i = \mathrm{Pool}_{\mu}\!\bigl\{ Q_j \mid j \in \mathcal{W}^{(\ell)}_i \bigr\},
   \quad                                                                 K^{(\ell)}_i = \mathrm{Pool}_{\mu}\!\bigl\{ K_j \mid j \in \mathcal{W}^{(\ell)}_i \bigr\},
   \quad                                                                  V^{(\ell)}_i = \mathrm{Pool}_{\mu}\!\bigl\{ V_j \mid j \in \mathcal{W}^{(\ell)}_i \bigr\},
   \label{eq:pool}
\end{equation}
with $\mathrm{Pool}_{\mu}$ denoting mean pooling over the window. Level~$0$ is the original full-resolution sequence ($\mathcal{W}^{(0)}_i = \{i\}$), and each subsequent level summarizes $p$ consecutive entries of the level below. We require $p^{L-1} \mid N$. Unlike prior hierarchical sparse designs (NSA, HISA, InfLLM-V2), which pool only the context side, Lighthouse applies $\mathrm{Pool}_{\mu}$ \emph{symmetrically} to all three projections. Symmetry buys two properties used in subsequent stages: a pooled query $Q^{(\ell)}_i$ and a pooled key $K^{(\ell)}_j$ live in the same representation space, and each pyramid entry is a coherent $(Q, K, V)$ triple summarizing the same $p^{\ell}$-token span. The total number of pyramid entries is $\sum_{\ell=0}^{L-1} N/p^{\ell} \le N \cdot p/(p-1)$, so pyramid construction costs $\Theta(N)$ time and memory.

\subsection{Scoring and Selection}
  \label{sec:method-scoring}
  Each pyramid entry receives two scalar scores --- one as a query, one as a key. At level $0$ we use per-head $\ell_2$ norms,
  \begin{equation}                                                                        s^{\mathrm{QK}}_{0, i} = \|Q_i\|_2,\qquad s^{\mathrm{KQ}}_{0, i} = \|K_i\|_2, \qquad i = 0, \dots, N-1,
  \label{eq:score-base}                                                                   \end{equation}  
  and at coarser levels we \emph{max-pool} from level $0$ rather than recomputing from pooled projections,                                                                     \begin{equation}                                                                        s^{\mathrm{QK}}_{\ell, i} = \max_{0 \le j < p^{\ell}} s^{\mathrm{QK}}_{0,\,i p^{\ell} +  j},\qquad s^{\mathrm{KQ}}_{\ell, i} = \max_{0 \le j < p^{\ell}} s^{\mathrm{KQ}}_{0,\,i p^{\ell} + j}.                                                                          \label{eq:score-pool}
  \end{equation}                                                                          Max-pooling lets a coarse span inherit the importance of its strongest token. Selection runs jointly over the concatenated $\mathrm{QK}$ and $\mathrm{KQ}$ streams across all levels via the chunked-bitonic kernel of Sec.~\ref{sec:topk}:
  
  \begin{equation}                                                                        \mathcal{I} \;=\; \mathrm{TopK}\!\left( \bigl\{ s^{\mathrm{QK}}_{\ell, i}, \; s^{\mathrm{KQ}}_{\ell, i} : (\ell, i) \in \mathcal{P} \bigr\}, \; k \right),            \label{eq:topk}                                                                         \end{equation}
  
  where $\mathcal{P}$ is the full set of pyramid indices. An entry chosen via its $\mathrm{KQ}$ score still enters the gather as its own $(Q^{(\ell)}_i, K^{(\ell)}_i, V^{(\ell)}_i)$ triple. The coarsest level is always retained in full --- it is cheap and guarantees at least one contributor at every base position; the remaining budget is spent on finer levels.

\subsection{Gathered-Sequence Attention}
  \label{sec:method-attn}
  Given $\mathcal{I}$, Lighthouse assembles a contiguous sub-sequence                                                                                                                      
  \begin{equation}                                                                                                                                                                         
  \widetilde{Q}_m = Q^{(\ell_m)}_{i_m},\quad \widetilde{K}_m = K^{(\ell_m)}_{i_m},\quad \widetilde{V}_m = V^{(\ell_m)}_{i_m},\qquad (\ell_m, i_m) \in \mathcal{I},\;\; m = 1, \dots, S,    
  \end{equation}                                                                                                                                                                           
  of length       
  \begin{equation}                                                                                                                                                                         
  S \;=\; \frac{N}{p^{L-1}} \;+\; (L-1)\,p\,k,
  \label{eq:S-size}                                                                                                                                                                        
  \end{equation}
  because the coarsest level contributes all $N/p^{L-1}$ entries while each of the remaining $L-1$ levels contributes at most $pk$ (the factor of $p$ is causal-boundary bookkeeping;      
  Sec.~\ref{sec:topk}). At $N=10^6, L=4, p=4, k=4096$, $S \approx 6.5\times 10^4 \ll N$. The sub-sequence is then attended to via stock SDPA or FlashAttention,
  
  \begin{equation}
  \widetilde{O} \;=\; \mathrm{Attn}(\widetilde{Q},\, \widetilde{K},\, \widetilde{V};\, \widetilde{M}),                                                                                     
  \label{eq:attn-sub}                                                                                                                                                                      
  \end{equation}
  where $\mathrm{Attn}(\cdot)$ is standard masked softmax attention. The causal mask $\widetilde{M}$ derives from the pyramid coordinates $(\ell_m, i_m)$ so each entry attends only to entries whose base positions are no greater than its own; the gather is topologically sorted, so $\widetilde{M}$ reduces to a standard $S{\times}S$ causal mask and Eq.~\eqref{eq:attn-sub} contains no sparse indexing.

  Due to the hierarchical decomposition, this gathering process guaranties that there are no "holes" or empty spaces in the sequence, which is especially important as we also compress queries Q; a hole could cause training instabilities as those missing tokens would be cut out during the forward pass and have no gradients during the backward pass. This is unlike asymmetrical methods that do not compress queries.
  
\subsection{Scatter-Back Reconstruction}                                                  \label{sec:method-scatter}
  The attention output $\widetilde{O} \in \mathbb{R}^{S \times d}$ is redistributed to the full $N$-token output $O$. A selected entry at level $\ell$, position $i$ summarized window     
  $\mathcal{W}^{(\ell)}_i$ during pooling but its output is written to a \emph{shifted} range                                                                                   \begin{equation}
  \mathcal{R}(\ell, i) \;=\; \bigl[\, i p^{\ell} + p^{\ell} - 1, \;\; i p^{\ell} + 2 p^{\ell} - 2 \, \bigr],                                                                 \label{eq:fanout}                                                                       \end{equation}                                                                          that starts at the last summarized token. The shift of $p^{\ell}-1$ preserves causality: a base position $j$ never receives a summary that contains its own future. Within a level, consecutive windows write to disjoint adjacent ranges; across levels, contributions are summed,                                                                                          
  \begin{equation}
  O_j \;=\; \sum_{m \,:\, j \in \mathcal{R}(\ell_m, i_m)} \widetilde{O}_m,                \label{eq:scatter}                                                                      \end{equation}                                                                          so the per-position fan-in is bounded by $L$ regardless of $k$.

    Similarly to the gathering pass, the scattering process also has no empty spaces. This final scattered sequence is fully dense, albeit a compressive approximation of full attention.

\section{Design Choices}                                         
\label{sec:design}                                                 
The Lighthouse pipeline of Sec.~\ref{sec:method} makes four design choices that distinguish it from prior selection-based sparse attention. First, $Q$ is pooled \emph{in lockstep} with $K, V$ instead of leaving queries dense as in NSA~\citep{yuan2025nsa}, HISA~\citep{zhao2026hisa}, and InfLLM-v2~\citep{zhao2026infllmv2}; this is the choice that turns the dense kernel call from $\mathcal{O}(NSd)$ to $\mathcal{O}(S^{2}d)$ at training time, and keeps pooled queries and pooled keys in the same representation space at every level. Second, the scorer is \emph{parameter-free} per-head $\ell_2$ norms of the layer's own $Q^{(\ell)}, K^{(\ell)}$ --- rather than a learned scoring head as in NSA~\citep{yuan2025nsa} or DSA~\citep{deepseek2025dsa}; this is the cheaper option and is strictly weaker than any attention- or QK-interaction-based scorer, so any positive result is a lower bound on what richer scorers can extract. The natural QK-interaction alternative we ablate against is a \emph{dilated softmax-attention} scorer that runs softmax attention over the pyramid with dilation factor $\delta$ at $\mathcal{O}(N^{2}d / \delta)$ per layer sub-quadratic but still super-linear in $N$, and an order of magnitude more expensive than the projection-norm scorer at long context (Sec.~\ref{sec:exp-ablations-summary}).

  Third, selection is \emph{decoupled} from attention: top-$K$ produces a contiguous, dense sub-sequence and attention is a stock SDPA or FlashAttention~\citep{dao2022flashattention,dao2023flashattention2} call on it, with no custom sparse-attention kernel coupling the two steps as in NSA~\citep{yuan2025nsa}, DSA~\citep{deepseek2025dsa}, or HISA~\citep{zhao2026hisa}. The same kernel runs at training and inference, and disabling selection cleanly recovers the dense baseline exactly the SDPA-resume test in Sec.~\ref{sec:exp-recoverability}. Fourth, we do \emph{not} make the top-$K$ differentiable: no straight-through estimator, no Gumbel
  softmax, no auxiliary scorer loss. Gradients flow only through the gathered $\widetilde{Q}, \widetilde{K}, \widetilde{V}$ into $W_Q, W_K, W_V$, so the projections learn to be useful \emph{when selected} rather than to game a learnable scorer. We motivate each choice and discuss alternatives in Appendix~\ref{sec:design-appendix}.

\section{Complexity Analysis and Kernel Design}                        
\label{sec:kernels}
Algorithm~\ref{alg:lighthouse} summarizes one Lighthouse attention layer as a sequence of GPU primitives. Most stages are standard operations executed via \texttt{torch.compile}'d PyTorch code; only the top-$k$ selection (stage~2c) and the scatter-back (stage~5) are custom kernels.

\begin{algorithm}[t]
  \DontPrintSemicolon
  \caption{Lighthouse attention (single layer, single head).}
  \label{alg:lighthouse}
  \KwIn{$X \in \mathbb{R}^{N \times d_\text{model}}$, projections $W_Q, W_K, W_V$, pyramid params $(L, p)$, budget $k$}
  \KwOut{$O \in \mathbb{R}^{N \times d}$} $Q, K, V \leftarrow X W_Q,\; X W_K,\; X W_V$                                                           \tcp*{projections (fused GEMM)}
  $\bigl\{(Q^{(\ell)}, K^{(\ell)}, V^{(\ell)})\bigr\}_{\ell=0}^{L-1}
  \leftarrow \mathrm{Pool}_{\mu}(Q, K, V)$
  \tcp*{pyramid (\texttt{view+mean})}
  $s^{\mathrm{QK}}_{\ell, i} \leftarrow \mathrm{Pool}_{\max}\!\bigl\{\|Q_j\|_2\bigr\},\;
  s^{\mathrm{KQ}}_{\ell, i} \leftarrow \mathrm{Pool}_{\max}\!\bigl\{\|K_j\|_2\bigr\}$
  \tcp*{scoring (\texttt{norm+max})}
  $\mathcal{I} \leftarrow \textsc{ChunkedBitonicTopK}(s^{\mathrm{QK}}, s^{\mathrm{KQ}}, k)$
  \tcp*{custom kernel (\S\ref{sec:topk})}
  $\widetilde{Q}, \widetilde{K}, \widetilde{V}
  \leftarrow \mathrm{Gather}\bigl(Q^{(\cdot)}, K^{(\cdot)}, V^{(\cdot)};\, \mathcal{I}\bigr)$
  \tcp*{\texttt{torch.gather}}
  $\widetilde{O} \leftarrow \mathrm{FlashAttention}(\widetilde{Q}, \widetilde{K}, \widetilde{V})$
  \tcp*{stock FA-3/FA-4}
  $O \leftarrow \textsc{ScatterBack}(\widetilde{O}, \mathcal{I})$
  \tcp*{custom kernel)}
  \Return $O$\;
\end{algorithm}

\subsection{Asymptotic Complexity}                              
  \label{sec:complexity}
  Table~\ref{tab:complexity} decomposes per-layer cost by stage. The only super-linear term in $N$ is the dense sub-sequence attention, $\Theta(S^2 d)$, with $S = N/p^{L-1} + (L-1)\,p\,k$ from Sec.~\ref{sec:method-attn}. Choosing $L = \log_p(N/k)$ balances the two terms in $S$, giving $S = \Theta(k \log_p(N/k))$ and an attention cost of $\Theta(k^2 \log^2 N \cdot d)$ --- polylogarithmic in $N$ at fixed $k$. Combined with the linear scoring and $\Theta(N \log k)$ selection passes, total per-layer compute is linear in $N$ up to a $\log k$ factor for bounded $k$. App.~\ref{app:complexity} derives this and compares against dense softmax, log-linear attention, and linear/SSM families.     
  
  \subsection{Kernel Design and Parallelism}
  \label{sec:kernel-summary}
  Of the seven stages in Algorithm~\ref{alg:lighthouse}, only top-$K$ and scatter-back are custom kernels in CUDA and triton; the rest reduce to PyTorch primitives that \texttt{torch.compile} fuses into single device passes. Our \emph{chunked-bitonic} top-$K$ partitions the score stream, maintains an in-register top-$m$ buffer per chunk via bitonic merge, and dispatches chunks as independent CTAs avoiding the shared-memory blow-up of textbook bitonic at $k{=}4096$ while producing a stratified selection that resists span collapse. Crucially, gather is decoupled from attention: where NSA~\citep{yuan2025nsa}, DSA~\citep{deepseek2025dsa}, HISA~\citep{zhao2026hisa}, and MoBA~\citep{lu2025moba} embed selection inside a custom sparse kernel, Lighthouse hands a contiguous dense sub-sequence to stock FlashAttention~\citep{shah2024flashattention3} --- making forward/backward bit-for-bit identical to a dense Transformer's, letting context parallelism rotate the gather through standard ring attention~\citep{liu2023ring} without any sparsity-aware collective, and enabling 1M-token training across 32 Blackwell GPUs (full details in App.~\ref{app:kernels}).

\section{Experiments}                                
  \label{sec:experiments}                                                                                                                                                                
  We evaluate Lighthouse along three axes: (1) \emph{recoverability}: whether lighthouse pretraining damages the model's ability to use full attention at inference Sec.~\ref{sec:exp-recoverability}); (2) \emph{design ablations and throughput} over the four knobs (scorer, $p$, $L$, $k$) and the resulting wall-clock cost (Sec.~\ref{sec:exp-ablations-summary}); and (3) \emph{scaling vs.\ dense attention} as a function of context length, including the long-context regime that requires context parallelism (Sec.~\ref{sec:exp-scaling}). All runs share the architecture and recipe of Sec.~\ref{sec:exp-setup}.
  
  \subsection{Experimental Setup}
  \label{sec:exp-setup}
  
  \paragraph{Architecture, data, optimizer.} A $530$M-parameter Llama-3-style decoder ($d_{\text{model}}{=}1024$, $30$ layers, $H{=}8$, head dim $128$, FFN $1536$, byte-level tokenizer). Layers $\{0,1,28,29\}$ retain dense SDPA --- PyTorch~$2.11.0$+\texttt{cu128}'s \texttt{torch.nn.attention.sdpa\_kernel} routed to cuDNN $9.19.0$ on CUDA $12.8$; the other 26 use Lighthouse with the same cuDNN-SDPA kernel as the inner attention call on the gathered sub-sequence. Training on C4 at sequence length $98{,}304$, global batch $32$, AdamW
  $2\!\times\!10^{-3}$, $\beta_1{=}0.9, \beta_2{=}0.95$, weight decay $0.1$, linear warmup over 2k steps, gradient-norm clip 1, bfloat16, FSDP only.                                       
  \paragraph{Two-stage recipe.} Stage~1 trains with Lighthouse; stage~2 resumes the stage-1 checkpoint under \emph{dense} SDPA (same cuDNN backend), with the same optimizer state and dataloader continuation. The total budget is held at $16{,}000$ steps ($\approx 50$~B tokens); we vary the stage-1 length to test sensitivity to the switch point.                       
  \paragraph{Hardware.} A single NVIDIA BGX 8$\times$B200 node is used for 98K-context runs; multi-node configurations are used with intra-node CP for 256K (Table~\ref{tab:ablation-overview}). We report training and validation loss, tokens/s per GPU in steady state, and total B200 hours.

  \subsection{SDPA Recoverability}   
  \label{sec:exp-recoverability}
  We test whether a hierarchical-trained Lighthouse model can be restored to dense attention by a brief continuation under stock SDPA. Holding the budget at $16{,}000$ steps ($\approx 50$\,B tokens), we vary the stage-1 length ($10$k / $11$k / $12$k) and resume the remainder under dense SDPA, against an full SDPA reference at matched architecture, data, and tokens (Table~\ref{tab:ablation-overview}, top block). At each resume the training loss transiently spikes ($1.12$--$1.57$) as the model is first run through attention it was not trained      
  against, then recovers within $\approx 1$--$1.5$k SDPA steps and crosses below the dense baseline; by step $16{,}000$ all three resume schedules match or beat dense-from-scratch        
  ($0.6980$--$0.7102$ vs.\ $0.7237$), with longer dense-resume tails giving lower final loss. Recovery is robust across resume points (the recipe doesn't pivot on a precise schedule),
  supporting our load-bearing claim that \emph{hierarchical training does not compromise the model's ability to use full attention at inference}, at no additional token cost over
  dense-from-scratch.

\begin{table}[!h]                                                                                  
  \centering                                                                                       
  \footnotesize
  \setlength{\tabcolsep}{5pt}
  \resizebox{\linewidth}{!}{%
  \begin{tabular}{@{}llcrrr|cc|c@{}}
  \toprule                                                                                                                                                                                 
  \textbf{Configuration} & \textbf{Scorer} & \textbf{Params} & \textbf{LH} & \textbf{Total} & \textbf{Total} & \textbf{B200-Hrs} & 
  \textbf{Tok/s (k)} & \textbf{Final Loss} \\ 
  
  \textbf{} & \textbf{} & \textbf{} & \textbf{Steps} & \textbf{Steps} & \textbf{Tokens} & \textbf{($\downarrow$)} & 
  \textbf{($\uparrow$)} & \textbf{($\downarrow$)} \\                                                                                            
  \midrule
  SDPA Baseline (ctx $=98k$)                          & ---     & 530M & ---  & 16k & 50.3B    & 303.2          & 45.6           & 0.7237 \\                                                                                                                                                                                   
  \midrule
  \multicolumn{9}{@{}p{12cm}}{\emph{SDPA recoverability} \newline \hspace{1em}($L{=}3,\ p{=}2,\ k{=}6144$, ctx $=98k$)} \\                                                                                 

  LH $\to$ SDPA (12k+4k)        & Dilated & 530M & 12k  & 16k & 50.3B    & \textbf{214.7} & 74.7           & 0.7102 \\                                              
  LH $\to$ SDPA (11k+5k)        & Dilated & 530M & 11k  & 16k & 50.3B    & 219.6          & \textbf{75.4}  & 0.7001 \\                                              
  LH $\to$ SDPA (10k+6k)        & Dilated & 530M & 10k  & 16k & 50.3B    & 228.0          & 75.0           & \textbf{0.6980} \\                                     
  \midrule                                                                                                                                                                              
  \multicolumn{9}{@{}p{12cm}}{\emph{Hyperparameter ablations} \newline \hspace{1em}(ctx $=98k$)} \\                                                                                                
  $L{=}3,\ p{=}2,\ k{=}1536$            & Dilated & 530M & 10k  & 16k & 50.3B    & 203.9          & 93.9           & \textbf{0.6825} \\                         
  &&&&&&&&\\
  $L{=}3,\ p{=}4,\ k{=}1536$            & Dilated & 530M & 10k  & 16k & 50.3B    & \textbf{197.2} & \textbf{99.5}  & 0.6881 \\                                              
  $L{=}3,\ p{=}8,\ k{=}1536$            & Dilated & 530M & 10k  & 16k & 50.3B    & 206.2          & 92.1           & 0.6828 \\                       
  &&&&&&&&\\
  $L{=}4,\ p{=}2,\ k{=}1536$            & Dilated & 530M & 10k  & 16k & 50.3B    & 200.2          & 96.4  & 0.6978 \\                                              
  $L{=}5,\ p{=}2,\ k{=}1536$            & Dilated & 530M & 10k  & 16k & 50.3B    & 201.5          & 96.3           & 0.6991 \\                       
  &&&&&&&&\\ 
  $L{=}3,\ p{=}2,\ k{=}2048$            & Dilated & 530M & 10k  & 16k & 50.3B    & 208.1          & 90.9           & 0.6880 \\                                                 
  $L{=}3,\ p{=}2,\ k{=}4096$            & Dilated & 530M & 10k  & 16k & 50.3B    & 215.7          & 83.5           &  0.6951 \\                                            
  \midrule                                                                                                                                                                                    
  \multicolumn{9}{@{}l}{\emph{CP training} ~~ ($L{=}3,\ p{=}4$)} \\                                                                                                         
  $k{=}1536$, ctx $=98k$, CP$=2$, DP$=4$    & Norm & 530M & 10k  & 16k & 100.7B    & 208.3   & 91.8  & 0.6903 \\                                                             
  $k{=}2048$, ctx $=98k$, CP$=2$, DP$=4$    & Norm & 530M & 10k  & 16k & 100.7B    & 210.9   & 89.2  & 0.6928 \\                                                             
  $k{=}4096$, ctx $=256k$, CP$=8$, DP$=1$   & Norm & 530M & 10k  & 16k & 1.07T & 1300.3 & 48.9  & \textbf{0.6721} \\                                                    
  \bottomrule                                                                                                                                                                              
  \end{tabular}}%
  \vspace{6pt}                                                                                                                                                                             
  \caption{\textbf{Lighthouse ablation summary.} 530M Llama-3, $16$k optimizer steps; \emph{LH Steps} run Lighthouse, \emph{SDPA Steps} run dense SDPA-resume on a single $8\!\times\!$B200
   node (the final block adds context parallelism). \textbf{B200-Hrs}: combined wall-clock $\times\,$8 GPUs. \textbf{Tok/s (k)}: Lighthouse-stage
  \texttt{throughput(tps)} from torchtitan, aggregated over all ranks (the baseline SDPA shows the throughput when training without LH). The full set of ablations is provided in Table~\ref{tab:ablation-overview-full}, Appendix~\ref{appendix:full_ablations}.}
  \label{tab:ablation-overview}
  \end{table}

  \subsection{Scaling Laws vs.\ Dense Attention}                                                                                                                                           
  \label{sec:exp-scaling}
\begin{figure}[h]
  \centering
  \begin{subfigure}{0.49\textwidth}
  \centering                                                                              \includegraphics[width=\linewidth]{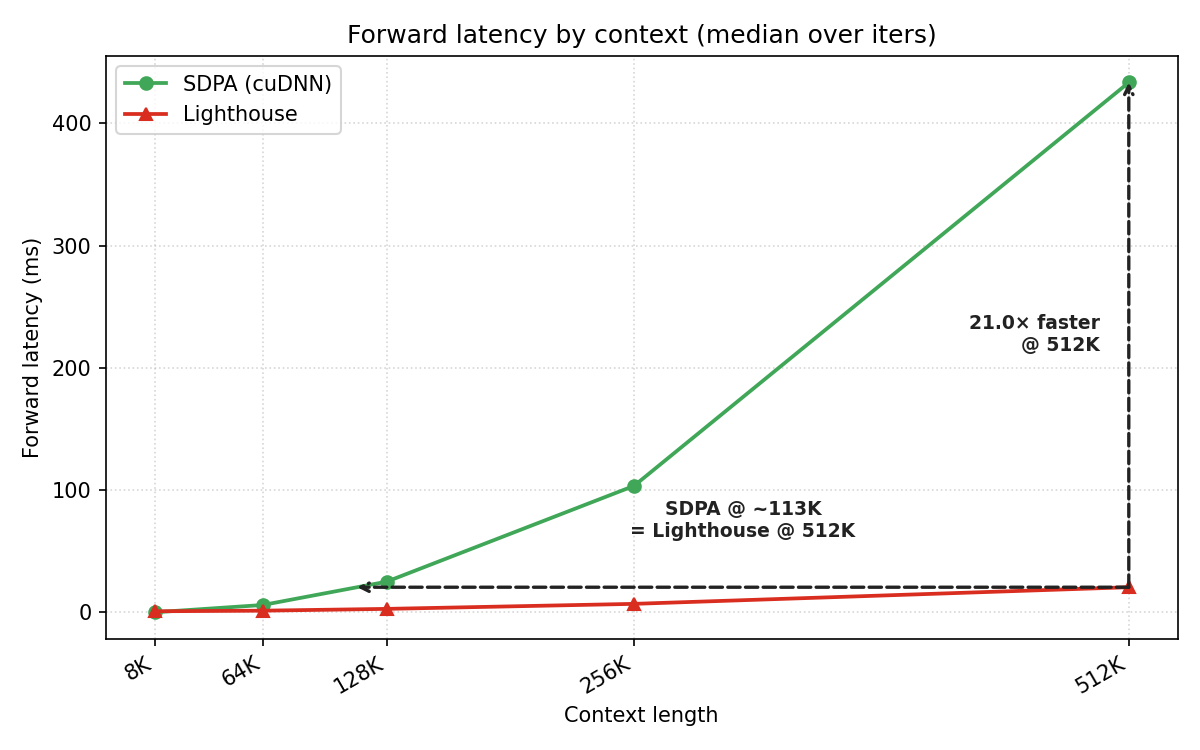}
  \caption{Forward Latency}                                               
  \label{fig:scaling-64x}                                                                 \end{subfigure}\hfill                                                                   \begin{subfigure}{0.49\textwidth}                                                       \centering                                                                              \includegraphics[width=\linewidth]{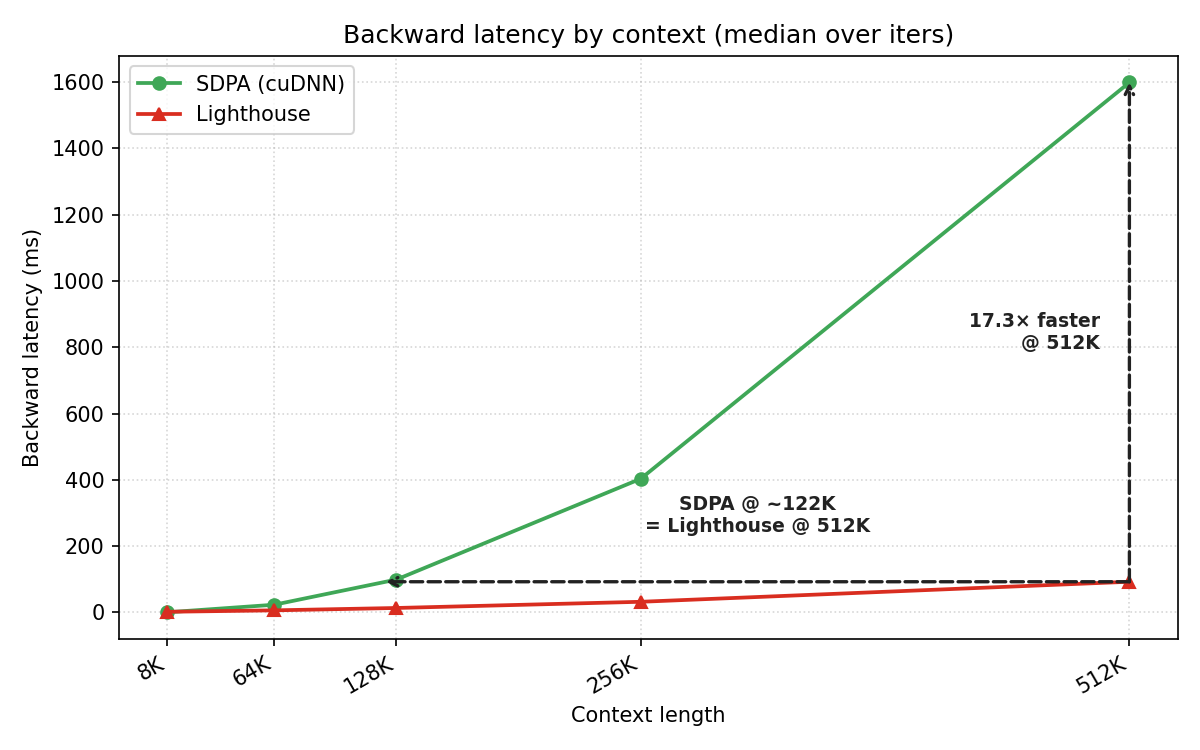}

  \caption{Backward Latency}                                               
  \label{fig:scaling-16x}
  \end{subfigure}                                                                         \caption{\textbf{Attention latency vs.\ context length} for SDPA (cuDNN) and Lighthouse (w/ cuDNN) on a single B200, $L{=}3,\ p{=}4$, sparsity $\approx 1{:}64$. SDPA scales as $\Theta(N^2 d)$;
  Lighthouse scales as $\Theta(S^2 d)$ with $S \ll N$. At $N{=}512$K, Lighthouse is $21{\times}$ faster on the forward pass and $17.3{\times}$ faster on the backward pass,      
  equivalently Lighthouse at 512k takes same runtime as if training SDPA at $\sim$113K / $\sim$122K context.}
  \label{fig:bench-fwd-bwd}                                                                     \end{figure}

  We benchmark single-layer attention latency on a single B200 for contexts from $8$K to $512$K (bf16, $B{=}1$, $H{=}8$, $d{=}128$, $L{=}3$, $p{=}4$, sparsity $1{:}64$, medians of $10$   
  steady-state iterations), comparing Lighthouse against cuDNN-backed SDPA. SDPA scales as $\Theta(N^2 d)$ while Lighthouse scales as $\Theta(S^2 d)$ with $S$ defined in                  
  Eq.~\ref{eq:S-size}, so the gap widens with $N$ (Fig.~\ref{fig:bench-fwd-bwd}). At $N{=}512$K, Lighthouse is $\boldsymbol{21\times}$ faster on the forward pass and                      
  $\boldsymbol{17.3\times}$ faster on forward$+$backward; equivalently, SDPA needs $\sim$$113$K (fwd) / $\sim$$122$K (fwd$+$bwd) of context to reach the runtime Lighthouse takes at       
  $512$K. Full-model training tells a similar story but requires care: with our $530$M-parameter architecture a single B200 OOMs beyond $\sim$$100$K on activations, gradients, and
  optimizer state regardless of attention method, so we implement context parallelism (Sec.~\ref{sec:cp}) where pyramid pooling, scoring, and top-$K$ run shard-locally and the gathered
  sub-sequence rotates through stock ring attention~\citep{liu2023ring} with no sparse-aware collectives. CP introduces a small ring-rotation overhead, costing $\sim$$10\%$ in per-rank
  throughput vs.\ the single-device extrapolation, but the Lighthouse-vs-SDPA speedup is preserved under matched CP geometry (Lighthouse-CP retains the same multiplicative gain over
  SDPA-CP that we see in the non-CP comparison), carrying the advantage cleanly to the $1$M-token / $32$-GPU regime.

\subsection{Design Ablations and Throughput}
\label{sec:exp-ablations-summary}

  We sweep four design axes (scorer variant, pooling factor $p$, number of levels $L$, top-$K$ budget $k$), each varied independently while the others stay at the defaults of             
  Sec.~\ref{sec:exp-setup}; comparisons use the post-resume training loss at step $16{,}000$. The full grid in Table~\ref{tab:ablation-overview} establishes three things. First, every
  Lighthouse configuration matches or beats the dense-SDPA-from-scratch baseline of $0.7237$, so recoverability is not specific to any one hyperparameter setting. 
  
  Second, the             
  projection-norm scorer is within $\sim$$0.01$ of dilated softmax in either direction (no uniform winner) but is parameter-free and roughly $9\%$ cheaper in B200-hours ($179.6$--$180.9$
  vs.\ $197.2$--$199.7$ at $L{=}3, p{=}4$). 
  
  Third, smaller $p$, shallower $L$, and smaller $k$ all help slightly: the lowest-loss cell across the grid is $L{=}3,\, p{=}2,\, k{=}1536$
  (dilated, loss $0.6825$), Pareto-best on every metric within its Top-$K$ block. The smaller-$k$ direction is the most counter-intuitive: loss decreases monotonically as $k$ shrinks over
   $\{1{,}536, 2{,}048, 3{,}072, 4{,}096\}$ ($0.6825 \to 0.6880 \to 0.6890 \to 0.6951$) before dipping again at $k{=}6{,}144$ ($0.6831$), plausibly because hierarchical selection regularises
  at our token budget; whether this reverses at substantially larger budgets is left to future work.

  The throughput story is consistent. Lighthouse stage-1 sustains $84$--$126$k tok/s/GPU across the ablation grid against $\sim$$46$k for dense SDPA, a roughly $2\times$ per-step         
  advantage that holds across selection budgets; the projection-norm scorer at $L{=}3,\, p{=}4,\, k{=}1536$ tops the range at $126$k by skipping the dilated-attention pass entirely. 
  
  End-to-end on the $10$k$+$$6$k recipe, total runtime ranges from $22.5$h ($179.6$ B200-h, norm $k{=}1536, p{=}4$) to $27.0$h ($215.7$ B200-h, $k{=}4096, p{=}2, L{=}3$) against $37.9$h ($303.2$ B200-h) for
  dense-SDPA-from-scratch on the same $16$k-step / $50.3$B-token budget: a $1.40\times$ to $1.69\times$ wall-clock speedup at matched or lower final loss. The saving comes entirely from
  stage-1; the SDPA-resume tail uses the same kernel as the baseline and matches its throughput. App.~\ref{app:ablations-throughput} gives the per-axis breakdowns, asymptotic-cost
  predictions, and per-stage timing decompositions.

\section{Conclusion}
\label{sec:discussion}
  We introduce Lighthouse Attention, a selection-based hierarchical attention for long-context pretraining that pools $Q, K, V$ \emph{symmetrically} across a multi-resolution pyramid and places 
  selection \emph{outside} the attention kernel, reducing the attention step to stock FlashAttention on a dense sub-sequence. The design is parameter-free, trains end-to-end with no      
  auxiliary losses or straight-through estimators, and inherits upstream FlashAttention improvements unchanged. A brief dense-SDPA resumption after Lighthouse pretraining matches or beats dense-from-scratch at matched tokens on training loss and long-context retrieval, with $1.4$--$1.7\times$ end-to-end speedups against cuDNN SDPA at $\ge 100$K context on B200 and clean scaling to $1$M tokens on multi-node Blackwell.
  
  \paragraph{Limitations.} Symmetric $Q/K/V$ pooling presumes all queries co-occur in one forward pass, which autoregressive decoding violates; we rely on dense-SDPA resumption for an    
  inference-ready model, and every downstream evaluation is run after that resume rather than on the hierarchical forward directly. The inner attention is $\Theta(S^2 d)$ on the gathered
  sub-sequence: sub-quadratic in $N$ at fixed $k$ but not strictly linear, so regimes where $k$ must scale with $N$ remain uncharacterised.                                                

  \paragraph{Future directions.} Swapping the dense-SDPA resume for an asymmetric sparse target (DSA, NSA, HISA, MoBA) would yield a natively serveable checkpoint; per-layer or per-head  
  adaptive $k$ may outperform a fixed budget; the multi-scale pyramid extends naturally to vision, audio, and video; and serving integration (continuous batching, speculative decoding,
  KV-cache management) is needed to translate the training speedups into deployment.

\bibliography{references}

\newpage

\appendix

\section{Ablations} \label{appendix:full_ablations}

\begin{table}[h]                                                                                  
  \centering                                                                                       
  \footnotesize
  \setlength{\tabcolsep}{5pt}
  \resizebox{\linewidth}{!}{%
  \begin{tabular}{@{}llcrrrccccc@{}}
  \toprule                                                                                                                                                                                 
  \textbf{Configuration} & \textbf{Scorer} & \textbf{Params} & \textbf{LH} & \textbf{Total} & \textbf{Total} & \textbf{B200-Hrs} & 
  \textbf{Tok/s (k)} & \textbf{Final Loss} \\ 
  
  \textbf{} & \textbf{} & \textbf{} & \textbf{Steps} & \textbf{Steps} & \textbf{Tokens} & \textbf{($\downarrow$)} & 
  \textbf{($\uparrow$)} & \textbf{($\downarrow$)} \\                                                                                            
  \midrule
  SDPA Baseline (ctx $=98k$)                          & ---     & 530M & ---  & 16k & 50.3B    & 303.2          & 45.6           & 0.7237 \\                                                                                                                                                                                   
  \midrule
  \multicolumn{9}{@{}p{12cm}}{\emph{SDPA recoverability} \newline \hspace{1em}($L{=}3,\ p{=}2,\ k{=}6144$, ctx $=98k$)} \\                                                                                 

  LH $\to$ SDPA (12k+4k)        & Dilated & 530M & 12k  & 16k & 50.3B    & \textbf{214.7} & 74.7           & 0.7102 \\                                              
  LH $\to$ SDPA (11k+5k)        & Dilated & 530M & 11k  & 16k & 50.3B    & 219.6          & \textbf{75.4}  & 0.7001 \\                                              
  LH $\to$ SDPA (10k+6k)        & Dilated & 530M & 10k  & 16k & 50.3B    & 228.0          & 75.0           & \textbf{0.6980} \\                                     
  \midrule                                                                                                                                                                                 
  \multicolumn{9}{@{}p{12cm}}{\emph{Scorer ablation} \newline \hspace{1em}($L{=}3,\ p{=}4$, ctx $=98k$)} \\                                                                                                
  $k{=}1536$                    & Dilated & 530M & 10k  & 16k & 50.3B    & 197.2          & 99.5           & \textbf{0.6881} \\                                     
  $k{=}1536$                    & Norm    & 530M & 10k  & 16k & 50.3B    & \textbf{179.6} & \textbf{126.0} & 0.6946 \\                                              
  $k{=}2048$                    & Dilated & 530M & 10k  & 16k & 50.3B    & 199.7          & 97.1           & 0.6969 \\                                              
  $k{=}2048$                    & Norm    & 530M & 10k  & 16k & 50.3B    & 180.9          & 122.4          & 0.6921 \\                                              
  \midrule                                                                                                                                                                                 
  \multicolumn{9}{@{}p{12cm}}{\emph{Pooling-factor ablation} \newline \hspace{1em}($L{=}3$, ctx $=98k$)} \\                                                                                                
  $p{=}2,\ k{=}1536$            & Dilated & 530M & 10k  & 16k & 50.3B    & 203.9          & 93.9           & \textbf{0.6825} \\                                     
  $p{=}4,\ k{=}1536$            & Dilated & 530M & 10k  & 16k & 50.3B    & \textbf{197.2} & \textbf{99.5}  & 0.6881 \\                                              
  $p{=}8,\ k{=}1536$            & Dilated & 530M & 10k  & 16k & 50.3B    & 206.2          & 92.1           & 0.6828 \\                                              
  $p{=}2,\ k{=}2048$            & Dilated & 530M & 10k  & 16k & 50.3B    & 208.1          & 90.9           & 0.6880 \\                                              
  $p{=}4,\ k{=}2048$            & Dilated & 530M & 10k  & 16k & 50.3B    & 199.7          & 97.1           & 0.6969 \\                                              
  \midrule                                                                                                                                                                                 
  \multicolumn{9}{@{}p{12cm}}{\emph{Number-of-levels ablation} \newline \hspace{1em}($p{=}2$, ctx $=98k$)} \\                                                                                              
  $L{=}3,\ k{=}1536$            & Dilated & 530M & 10k  & 16k & 50.3B    & 203.9          & 93.9           & \textbf{0.6825} \\                                     
  $L{=}4,\ k{=}1536$            & Dilated & 530M & 10k  & 16k & 50.3B    & \textbf{200.2} & \textbf{96.4}  & 0.6978 \\                                              
  $L{=}5,\ k{=}1536$            & Dilated & 530M & 10k  & 16k & 50.3B    & 201.5          & 96.3           & 0.6991 \\                                              
  $L{=}3,\ k{=}2048$            & Dilated & 530M & 10k  & 16k & 50.3B    & 208.1          & 90.9           & 0.6880 \\                                              
  $L{=}4,\ k{=}2048$            & Dilated & 530M & 10k  & 16k & 50.3B    & 202.2          & 94.5           & 0.6983 \\                                              
  $L{=}5,\ k{=}2048$            & Dilated & 530M & 10k  & 16k & 50.3B    & 206.5          & 92.3           & 0.7043 \\                                              
  \midrule                                                                                                                                                                                 
  \multicolumn{9}{@{}p{12cm}}{\emph{Top-$K$ budget ablation} \newline \hspace{1em}($L{=}3,\ p{=}2$, ctx $=98k$)} \\                                                                                        
  $k{=}1536$                 & Dilated & 530M & 10k  & 16k & 50.3B    & \textbf{203.9} & \textbf{93.9}  & \textbf{0.6825} \\                                     
  $k{=}2048$                 & Dilated & 530M & 10k  & 16k & 50.3B    & 208.1          & 90.9           & 0.6880 \\                                              
  $k{=}3072$                 & Dilated & 530M & 10k  & 16k & 50.3B    & 214.9          & 86.1           &  0.6890 \\                                              
  $k{=}4096$                 & Dilated & 530M & 10k  & 16k & 50.3B    & 215.7          & 83.5           &  0.6951 \\                                              
  $k{=}6144$                 & Dilated & 530M & 10k  & 16k & 50.3B    & 208.1          & 88.3           &  0.6831 \\                                              
  \midrule                                                                                                                                                                                 
  \multicolumn{9}{@{}l}{\emph{CP training} ~~ ($L{=}3,\ p{=}4$)} \\                                                                                                         
  $k{=}1536$, ctx $=98k$, CP$=2$, DP$=4$    & Norm & 530M & 10k  & 16k & 100.7B    & 208.3   & 91.8  & 0.6903 \\                                                             
  $k{=}2048$, ctx $=98k$, CP$=2$, DP$=4$    & Norm & 530M & 10k  & 16k & 100.7B    & 210.9   & 89.2  & 0.6928 \\                                                             
  $k{=}4096$, ctx $=256k$, CP$=8$, DP$=1$   & Norm & 530M & 10k  & 16k & 1.07T & 1300.3 & 48.9  & \textbf{0.6721} \\                                                    
  \bottomrule                                                                                                                                                                              
  \end{tabular}}%
  \vspace{6pt}                                                                                                                                                                             
  \caption{\textbf{Full Lighthouse ablation summary.} 530M Llama-3, $16$k optimizer steps; \emph{LH Steps} run Lighthouse, \emph{SDPA Steps} run dense SDPA-resume on a single $8\!\times\!$B200
   node (the final block adds context parallelism). \textbf{B200-Hrs}: combined wall-clock $\times\,$8 GPUs. \textbf{Tok/s (k)}: Lighthouse-stage \texttt{throughput(tps)} from torchtitan, aggregated over 8 ranks (the Dense-SDPA-from-scratch row has no LH stage and reports its dense values). \textbf{Final Loss}: training loss at step $16{,}000$. Bold marks the per-block best on each metric (skipped for
  throughput in the CP block since context length varies inside it).}
  \label{tab:ablation-overview-full}
  \end{table}

\newpage

\section{Complexity Derivation}
\label{app:complexity}

\begin{table}[h]                                                                        \centering
  \small                                                                                    \begin{tabular}{@{}lll@{}}
    \toprule
    Stage & Primitive & Cost \\
    \midrule                                                                                                                                                                                 
    Projections $Q, K, V$        & GEMM                  & $\Theta(N\, d_\text{model}\, d)$ \\
    Pyramid pool                 & \texttt{view+mean}    & $\Theta(N\, d)$ \\                                                                                                                
    Scoring (norms, max-pool)    & \texttt{norm+max}     & $\Theta(N\, d)$ \\                                                                                                                
    Top-$k$ selection            & chunked bitonic       & $\Theta(N \log k)$ \\                                                                                                             
    Gather to sub-sequence       & \texttt{torch.gather} & $\Theta(S\, d)$ \\                                                                                                                
    Dense sub-sequence attention & FlashAttention        & $\Theta(S^2\, d)$ \\                                                                                                              
    Scatter-back                 & custom atomic         & $\Theta(N\, d)$ \\                                                                                                                
  \bottomrule
  \end{tabular}
  \vspace{4pt}
  \caption{Per-layer complexity of a Lighthouse attention layer, with $S = N/p^{L-1} + (L-1)\,p\,k$.}
  \label{tab:complexity}
\end{table}

We derive the per-layer compute complexity of Lighthouse and show that, at bounded selection budget $k$, total compute is linear in the sequence length $T$.

\paragraph{Setup.} Lighthouse with sequence length $T$, pooling factor $p$, $L$ pyramid levels, top-$k$ budget $k$, and head dimension $d$.

\paragraph{Per-stage cost.} Table~\ref{tab:complexity} decomposes one layer into its stages.

\paragraph{Sub-sequence size.} By construction the gathered sub-sequence has size
\begin{equation}
    S \;=\; \frac{T}{p^{L-1}} \;+\; (L-1)\,p\,k,
      \label{eq:S-base}
\end{equation} where the first term is the coarsest level (kept whole) and the second is the contribution of the finer $L-1$ levels (each of which contributes at most $p\,k$ entries).

\paragraph{Choice of $L$.} Setting $L = \log_p(T/k)$ gives $p^{L} = T/k$ and therefore $p^{L-1} = T/(p\,k)$. Substituting into Eq.~\eqref{eq:S-base},
\[
    \frac{T}{p^{L-1}} \;=\; p\,k,
\] and the two terms combine into
\begin{equation}
    S \;=\; p\,k + (L-1)\,p\,k \;=\; p\,k \cdot L \;=\; p\,k \cdot \log_p(T/k) \;=\; \Theta(k \log T)
    \label{eq:S-asymptotic}
\end{equation}
(treating $p$ as constant).

\paragraph{Attention cost in terms of $T$.} The dense FlashAttention call on $S$ tokens costs $\Theta(S^{2} \cdot d)$. Substituting Eq.~\eqref{eq:S-asymptotic},
\[
    S^{2} \cdot d \;=\; \Theta\!\left(k^{2} \log^{2} T \cdot d\right),
\]
which is polylogarithmic in $T$ for bounded $k$.

\paragraph{Total per-layer compute.} Summing the contributions in Table~\ref{tab:complexity} after substituting $S = \Theta(k \log T)$,  
  \[
      T_{\text{layer}}                                                                                      
      \;=\;       
      \Theta(T \cdot d)                                                                                     
      \;+\;
      \Theta(T \log k)                                                                                      
      \;+\;       
      \Theta\!\left(k^{2} \log^{2} T \cdot d\right).
  \]
  
For bounded $k$, the polylog term is sub-linear in $T$ and the $\Theta(T \cdot d)$ term from the projection, pooling, scoring, and scatter stages dominates. Therefore

\begin{equation}
    \boxed{\,T_{\text{layer}} \;=\; \Theta(T \cdot d) \;\;\text{at bounded }k.\,}
    \label{eq:total-cost}
\end{equation}

\paragraph{Two distinct quantities.} We emphasize that two ``log'' factors appear at different points and should not be conflated:

\begin{itemize}[noitemsep, topsep=2pt]
    \item Sub-sequence size: $S = \Theta(k \log T)$ the size of the data tensor passed to FlashAttention. 
    \item Per-layer compute: $\Theta(T \cdot d)$ the total flops to run one layer.
\end{itemize}

The logarithmic factor lives in $S$; the total compute is linear in $T$ because only $S$ tokens (not $T$) attend to one another, while the linear-cost stages remain linear in $T$.

\paragraph{Comparison to other attention families.} Table~\ref{tab:complexity-cmp} places Lighthouse alongside dense softmax, log-linear attention, and linear-attention / SSM families.

\begin{table}[h]
    \centering
    \small
    \begin{tabular}{@{}ll@{}}
        \toprule
        Method & Per-layer compute \\
        \midrule
        Dense softmax                            & $\Theta(T^{2} \cdot d)$ \\
        Log-Linear Attention~\citep{guo2025log}  & $\Theta(T \log T \cdot d)$ \\
        \textbf{Lighthouse} (bounded $k$)        & $\boldsymbol{\Theta(T \cdot d)}$ \\
        Linear attention / SSMs (fixed budget)   & $\Theta(T \cdot d)$ \\
        \bottomrule
    \end{tabular}
    \vspace{6pt}
    \caption{Per-layer compute across attention families. Lighthouse and linear/SSM families share the same asymptotic class; log-linear attention is strictly worse asymptotically; dense softmax is quadratic.}
    \label{tab:complexity-cmp}
\end{table}

 \section{Design Choices (extended)}
  \label{sec:design-appendix}
  The Lighthouse pipeline of Sec.~\ref{sec:method} admits several non-trivial design decisions whose rationale is not obvious from the equations alone.
  
  \subsection{Symmetric Q/K/V Pooling}                                                                                                         
  \label{sec:sym-pool}
  Prior hierarchical sparse designs~\citep{yuan2025nsa,zhao2026hisa,zhao2026infllmv2} compress only the key--value side and leave queries at
  full resolution. This is natural for inference, where autoregressive decoding presents one query at a time. Training, however, exposes every 
  query in parallel: we can compress the query side too, then recover dense behavior at inference by briefly resuming with stock SDPA~\citep{dao2022flashattention}. Lighthouse pools $Q$ in lockstep with $K, V$ at every level, reducing the dense     
  FlashAttention call from $\mathcal{O}(N\,S\,d)$ to $\mathcal{O}(S^{2}\,d)$ and yielding coherent $(Q^{(\ell)}, K^{(\ell)}, V^{(\ell)})$ triples that share a representation space across levels --- pooled queries route to pooled keys, producing summary--summary interactions an asymmetric pyramid cannot express. Sec.~\ref{sec:exp-recoverability} verifies the symmetric design is invariant under the recovery test.

  \subsection{Parameter-Free Scoring}
  \label{sec:param-free-scoring}
  The scoring functional has the widest design space; two natural candidates are (a) a dilated softmax attention over the pyramid (most faithful to ``what would softmax do,'' as used by the learned selectors in NSA~\citep{yuan2025nsa} and DSA~\citep{deepseek2025dsa}, but $\mathcal{O}(N^{2}/r)$) and (b) the per-head $\ell_2$ norms of the layer's own projections, $\|Q^{(\ell)}_i\|_2$ and $\|K^{(\ell)}_i\|_2$ (no parameters, no Q--K interaction, $\Theta(N)$). We adopt the projection-norm scorer. It is the cheaper of the two and the more \emph{conservative} benchmark: a dilated-attention scorer strictly adds Q--K interaction information and can only help. Any positive result with projection-norms is therefore a lower bound on what Lighthouse can deliver, and our ablations in Appendix.~\ref{appendix:full_ablations} confirm the dilated scorer matches projection-norms within noise --- evidence that the \emph{selection structure}, not the scoring function, drives
  the long-context behavior.                                                                    
                  
  \subsection{Selection--Attention Decoupling}
  \label{sec:decoupling}
  Every competing selection method we are aware of fuses its selection machinery into the attention kernel: NSA~\citep{yuan2025nsa}, DSA~\citep{deepseek2025dsa}, and HISA~\citep{zhao2026hisa} each ship a custom sparse-attention kernel that reads indices from a selection step. Lighthouse places that interface \emph{outside} the kernel: selection produces a dense, contiguous sub-sequence and attention is a stock FlashAttention~\citep{dao2022flashattention,dao2023flashattention2,shah2024flashattention3} call on it. Two consequences follow: the same kernel runs at training and inference, so there is no train-vs-serve kernel divergence; and correctness of the attention step can be checked against the dense baseline by running attention with selection disabled, which is exactly what the SDPA-resume evaluation in
  Sec.~\ref{sec:exp-recoverability} exercises.

  \subsection{Gradient Flow}
  \label{sec:grad-flow}
  The top-$K$ step is discrete and we do not approximate it with a straight-through estimator~\citep{bengio2013ste} or Gumbel softmax~\citep{jang2017gumbel,maddison2017concrete}. Gradients flow back from the loss through the scatter, FlashAttention~\citep{dao2022flashattention}, and gather into $W_Q, W_K, W_V$ via the differentiable values $\widetilde{Q}, \widetilde{K}, \widetilde{V}$; selection indices and the scoring functional carry no gradient. The projections therefore learn to produce values that are \emph{useful when selected} rather than scores that are good at selecting --- avoiding the optimization pathologies (scorer collapse, scorer--attention misalignment, auxiliary-loss tuning) that learnable selectors~\citep{yuan2025nsa,deepseek2025dsa} are prone to.

  \section{Kernel Design and Parallelism (extended)}
  \label{app:kernels}
  \subsection{Pyramid and Scoring}                               \label{sec:kernels-pyramid}
  The pyramid and scoring stages are not custom kernels: pooling is a \texttt{view + mean}, scoring is a per-token \texttt{norm} followed by \texttt{view + max} per coarser level. All ops are pointwise or reshape-plus-reduce and are fused by \texttt{torch.compile} into a single device pass.
  
  \subsection{Chunked-Bitonic Top-$k$ Kernel}                    \label{sec:topk}
  Selecting $k$ pyramid entries out of $\Theta(N)$ candidates is the first stage that warrants a custom kernel. A textbook bitonic top-$k$ sorts the entire score stream in shared memory or registers, which fails at our budgets: $k = 4096$ over a pyramid of $\approx 2N$ entries cannot fit in a single thread block, and a global sort serializes across the sequence. Lighthouse instead uses a \emph{chunked-bitonic} design: the score stream is partitioned into fixed-size chunks of $N_\text{chunk}=2048$ scores, each chunk maintains a running top-$m$ buffer ($m=128$) updated through an in-register bitonic merge, and the $N/N_\text{chunk}$ chunks dispatch as independent CTAs. No thread block ever holds more than $m$ scores; the work is fully parallel.

  The concatenated indices arrive in score-sorted order within each chunk and chunk-major order across chunks --- not the causal order the attention step requires. We apply a single \texttt{torch.sort} pass to re-order them by pyramid position, after which the gathered $\widetilde{Q}, \widetilde{K}, \widetilde{V}$ form a contiguous, causally consistent sub-sequence indistinguishable in shape from a standard 
  dense FlashAttention input. This is what makes the rest of the pipeline kernel-agnostic.

  This design does not produce the same index set as a theoretical global top-$k$: if the $k$ globally highest-scoring entries cluster in one chunk, some are replaced by lower-scoring entries from other chunks. Read positively, this is a \emph{stratified} top-$k$ that guarantees every region of the sequence contributes some tokens, which empirically yields more balanced attention coverage than strict global top-$k$   
  and avoids selection collapse onto a narrow span.

  \subsection{Gather and FlashAttention Dispatch}                                                                               \label{sec:kernels-attn}
  Once $\mathcal{I}$ is produced, gather is a stock \texttt{torch.gather} followed by a single FlashAttention call on the gathered tensors. This is where placing selection \emph{outside} the attention kernel pays off. Prior selection-based methods~\citep{yuan2025nsa,deepseek2025dsa,zhao2026hisa,lu2025moba} embed the selection--attention interface inside the kernel: each tile reads a sparse index list and performs an indirect KV load, which forces (i) a custom forward kernel per GPU architecture, (ii) a matching custom backward that inverts the hierarchical pattern, and (iii) ongoing tile/schedule re-tuning as hardware evolves. Lighthouse's dense sub-sequence design sidesteps all three: forward and backward are bit-for-bit identical to a standard dense Transformer's.

  \subsection{Context-Parallel Execution}
  \label{sec:cp}
  At sequence lengths beyond 128K we train with context parallelism across $W$ devices, each rank holding a contiguous slice and using standard ring attention. Lighthouse extends cleanly without custom collectives because its pre-attention primitives are local: \emph{(i)} the coarsest pool window $p^{L-1}$ (e.g.\ 64) is orders of magnitude smaller than the shard size ($N/W = 128\text{K}$ at $N{=}1\text{M}, W{=}8$), so pooling and scoring need no inter-rank communication; \emph{(ii)} each rank runs the chunked-bitonic top-$k$ on its own pyramid, producing $\mathcal{I}_r \subseteq \mathcal{P}_r$ from tokens it already owns; \emph{(iii)} the gathered sub-sequence is dense, so FlashAttention runs under standard ring attention~\citep{liu2023ring} --- KV shards rotate through the ring as in a fully dense long-context run, and each rank's queries see the cross-shard context selected by every other rank's Lighthouse pipeline. This last property is only
  possible because Lighthouse's selection output is a contiguous tensor; sparse-selection kernels cannot express ring rotation without engineering specific to the sparse layout. The combined design supports 1M-token pretraining across 32 Blackwell GPUs (4 nodes × 8 GPUs, CP degree 8) with no changes to the attention kernel itself.

 \section{Design Ablations and Throughput (extended)}                                               \label{app:ablations-throughput}

 This appendix gives the per-axis ablation results in detail, the per-stage throughput decomposition, and the asymptotic-cost predictions backing each design choice. All loss numbers are the step-$16{,}000$ post-resume training loss of the two-stage Lighthouse-then-SDPA recipe (Sec.~\ref{sec:exp-recoverability}); throughput numbers are aggregated over $8$ ranks of one
  B200 node at $98{,}304$-token context (matching Table~\ref{tab:ablation-overview}).                                                                                                      
                  
  \subsection{Scorer Variants}
  \label{app:exp-abl-scorer}
  We compare the parameter-free projection-norm scorer against the dilated softmax-attention scorer at fixed $L{=}3,\, p{=}4$. At $k{=}1536$, dilated reaches $0.6881$ while
  projection-norm reaches $0.6946$; at $k{=}2048$ the order reverses, with $0.6969$ for dilated and $0.6921$ for projection-norm. The two are within $\sim$$0.01$ of each other in both    
  directions and neither is uniformly better. Combined with projection-norm's lower compute cost ($179.6$--$180.9$ B200-h vs.\ $197.2$--$199.7$ for dilated, an $\sim$$9\%$ saving) and
  absence of additional learnable parameters (Sec.~\ref{sec:param-free-scoring}), projection-norm is the throughput-sensitive default; dilated remains a defensible alternative when the   
  lowest absolute loss matters.

  \subsection{Pooling Factor}
  \label{app:exp-abl-pool}
  Holding $L{=}3$ and varying $p \in \{2, 4, 8\}$ at $k{=}1536$ gives final losses of $0.6825$, $0.6881$, $0.6828$ respectively; at $k{=}2048$, $p{=}2$ and $p{=}4$ give $0.6880$ and
  $0.6969$. Pooling factor has a small effect: $p{=}2$ and $p{=}8$ are essentially indistinguishable, with $p{=}4$ marginally worse by $0.005$--$0.015$. We adopt $p{=}2$ as the default; $p{=}4$ paired with the projection-norm scorer is the wall-clock-favoured alternative.
  
  \subsection{Number of Levels}
  \label{app:exp-abl-levels}
  Holding $p{=}2$ and varying $L \in \{3, 4, 5\}$ produces a monotonically increasing final loss: at $k{=}1536$, $L{=}3$ gives $0.6825$, $L{=}4$ gives $0.6978$, and $L{=}5$ gives
  $0.6991$; the ordering carries to $k{=}2048$ ($0.6880 \to 0.6983 \to 0.7043$). Deeper pyramids spread the same selection budget over more coarse levels, leaving fewer entries at the    
  finest level where the model is most sensitive. $L{=}3$ is best across both selection budgets and we adopt it as the default.
  
  \subsection{Top-$K$ Budget}
  \label{app:exp-abl-topk}
  Holding $L{=}3,\, p{=}2$ and varying $k \in \{1{,}536, 2{,}048, 3{,}072, 4{,}096, 6{,}144\}$ gives final losses of $0.6825,\, 0.6880,\, 0.6890,\, 0.6951,\, 0.6831$. The trend is
  monotonic over $k \le 4096$ and dips back at $k{=}6144$. A larger selection budget does \emph{not} translate to lower post-resume loss within the range we tested. Sparser configurations
   may regularise against our relatively small training-token budget; investigating whether this trend reverses at much larger budgets is left to future work. The practical implication is
   that $k{=}1536$ is preferred from both a loss and a wall-clock standpoint.                                                               
  \begin{figure}[h]
  \centering
  \includegraphics[width=\linewidth]{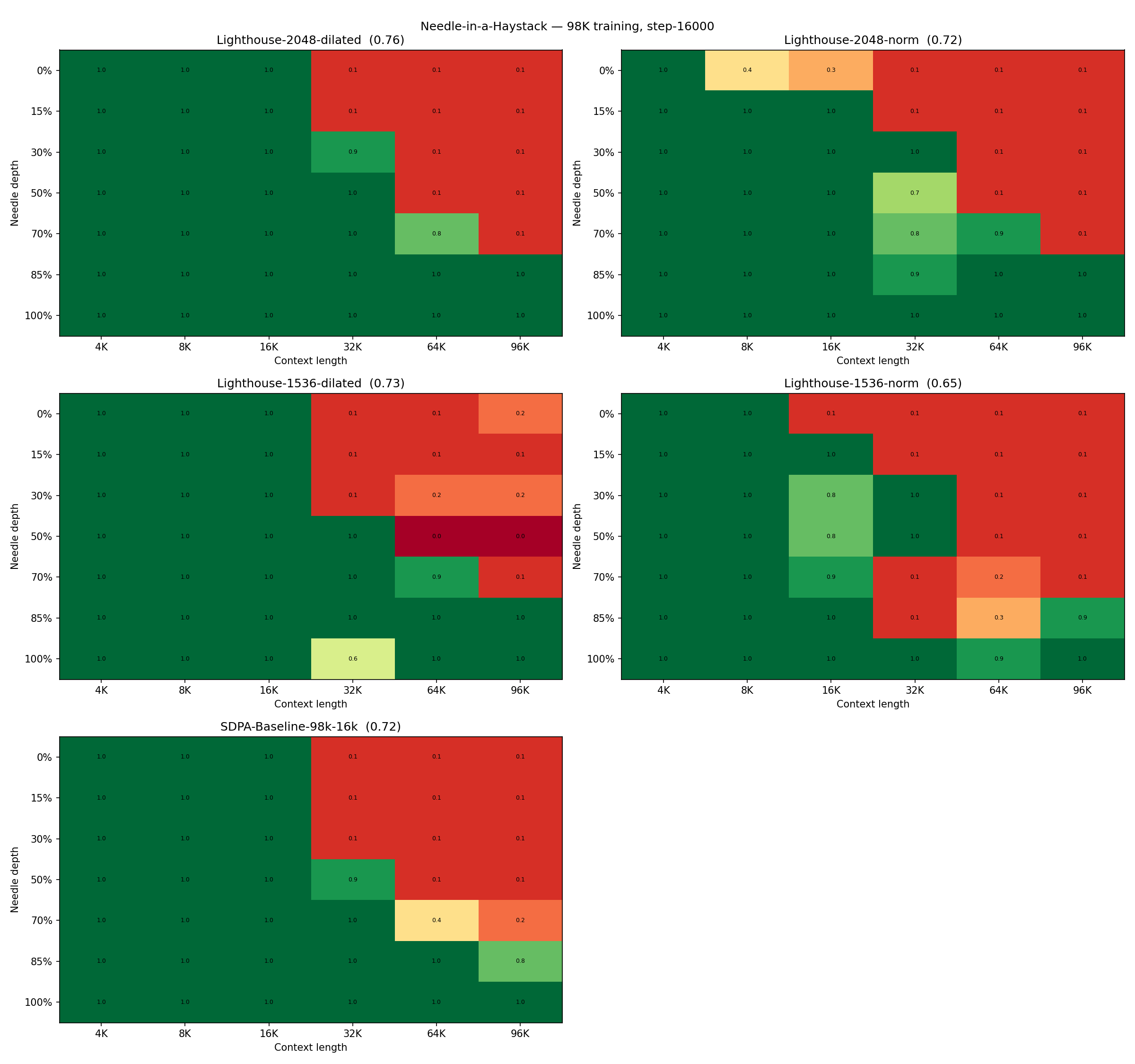}
  \caption{\textbf{Needle-in-a-Haystack at $98$K training, step $16{,}000$.} Four Lighthouse $\to$ SDPA configurations (varying $k \in \{1536, 2048\}$ and scorer $\in \{$dilated, norm$\}$ at $L{=}3,\, p{=}4$) and the dense SDPA-from-scratch baseline (bottom). Each cell is the mean retrieval rate over $10$ single-digit passkeys at the given (context, depth); the         
  per-panel mean is shown in each title. Random chance is $10\%$.}                                                                                                                         
  \label{fig:niah-grid}                                                                                                                                                                    
  \end{figure}                                                
                  
  \subsection{Throughput Decomposition}
  \label{app:exp-throughput}

  \paragraph{Stage-1 throughput.} The recoverability runs ($k{=}6144,\, L{=}3,\, p{=}2$, dilated) sustain $74.7$--$75.4$k tok/s/GPU. Across the ablation grid (varying $k,p,L$ at the $98$K
   context) the stage-1 throughput range is $83.5$--$126.0$k tok/s/GPU, vs.\ $\sim$$46$k for dense SDPA. The intra-grid trends follow the $S = N/p^{L-1} + (L-1)pk$ prediction: raising $p$
   from $2$ to $4$ at $k{=}2048, L{=}3$ lifts throughput from $90.9$ to $97.1$k; raising $L$ from $3$ to $4$ at $k{=}2048, p{=}2$ lifts it from $90.9$ to $94.5$k (deeper pyramid spreads  
  the budget across more coarse levels). Swapping the dilated scorer for projection-norm at $L{=}3,\, p{=}4,\, k{=}1536$ accelerates stage-1 from $99.5$ to $126.0$k tok/s/GPU because the
  scorer no longer runs an attention pass over the dilated pyramid.

  \paragraph{End-to-end speedup.} Aggregating both stages of the $10$k$+$$6$k recipe across the ablation grid, end-to-end runtime ranges from $22.5$h ($179.6$ B200-h; norm $k{=}1536,\,   
  p{=}4,\, L{=}3$) to $27.0$h ($215.7$ B200-h; $k{=}4096,\, p{=}2,\, L{=}3$) against $37.9$h ($303.2$ B200-h) for dense-SDPA-from-scratch on the same $16{,}000$-step / $50.3$B-token
  budget: a $1.40\times$ to $1.69\times$ wall-clock speedup at matched or lower final loss. The saving comes entirely from stage-1; the SDPA-resume tail runs the same kernel as the       
  baseline and matches its throughput.
  
  \section{Long-Context Retrieval (NIAH)}
  \label{app:niah}
  At 530M parameters and only $16{,}000$ training steps ($\sim$50.3B tokens), full prose Needle-in-a-Haystack scores near-zero across the
  board, so we adopt a simplified single-digit variant that isolates the retrieval signal. To complement the loss-based recoverability         
  evaluation in Sec.~\ref{sec:exp-recoverability}, we run this test over five step-$16{,}000$ checkpoints: four Lighthouse $\to$ SDPA two-stage runs (varying $k\in\{1536, 2048\}$ and scorer $\in\{$dilated, norm$\}$ at $L{=}3,\, p{=}4$) and the dense-SDPA-from-scratch baseline at     
  matched compute and tokens. Inference uses dense SDPA in every case (Sec.~\ref{sec:exp-setup}).                                                                                              
                  
  \paragraph{Protocol.} A single passkey digit (one of $\{0, 1, \dots, 9\}$) is hidden in random alphanumeric filler at depths $\{0, 15, 30, 50, 70, 85, 100\}\%$ across context lengths   
  $\{4, 8, 16, 32, 64, 96\}$K. For each cell we run one forward pass over the full prompt and take an argmax restricted to the 10 digit tokens at the last position. We average the 0/1
  score over the full digit sweep, so each cell reports the mean retrieval rate over $10$ trials; random chance is $10\%$.                                                                 
                  
  \paragraph{Findings.} Three of the four Lighthouse runs are at or above the SDPA-from-scratch baseline (mean retrieval $0.72$): $k{=}2048$ dilated wins overall at $0.76$, $k{=}1536$    
  dilated reaches $0.73$, and $k{=}2048$ norm matches the baseline at $0.72$; only $k{=}1536$ norm dips, to $0.65$ (Fig.~\ref{fig:niah-grid}). Two patterns emerge. First, larger $k$ is
  the dominant axis: $k{=}2048 \ge k{=}1536$ for both scorers, with a gap of $0.03$ (dilated) or $0.07$ (norm). Second, the norm scorer hurts retrieval more than it hurts training loss:  
  at fixed $k$, switching dilated to norm costs $0.04$ at $k{=}2048$ and $0.08$ at $k{=}1536$, the largest single-axis gaps in the grid. Combined with the loss-side finding that smaller
  $k$ regularises better, the right default depends on whether the downstream task is loss- or retrieval-driven.

\end{document}